\documentclass[11pt, a4paper, copyright]{simular}

\usepackage{microtype}
\usepackage{hyperref}
\usepackage{url}
\usepackage{booktabs}
\usepackage{tcolorbox}
\usepackage{subfigure}

\usepackage{lineno}

\usepackage{listings}
\usepackage{xcolor}
\lstdefinestyle{prompt}{
  basicstyle=\ttfamily\small,
  numbers=left,
  numberstyle=\tiny\color{gray},
  numbersep=6pt,
  frame=single,
  framerule=0.4pt,
  rulecolor=\color{gray!50},
  backgroundcolor=\color{gray!3},
  breaklines=true,
  breakatwhitespace=false,
  columns=fullflexible,
  showstringspaces=false,
  upquote=true
}

\usepackage{amsmath}
\usepackage{amsthm}
\usepackage{amssymb}
\usepackage{stmaryrd}
\usepackage{latexsym}
\usepackage{mathtools}
\usepackage{cleveref}
\usepackage{enumitem}
\definecolor{darkblue}{rgb}{0, 0, 0.5}
\hypersetup{colorlinks=true, citecolor=darkblue, linkcolor=darkblue, urlcolor=darkblue}

\newcommand{\passk}[1]{{Pass\textasciicircum#1}}

\newif\ifcomments
\commentsfalse  

\ifcomments
  \newcommand{\jc}[1]{{\color{red} jc: #1}}
  \newcommand{\eric}[1]{{\color{blue} Eric: #1}}
  \newcommand{\richard}[1]{{\color{green} Richard: #1}}
  \newcommand{\vincent}[1]{{\color{teal} Vincent: #1}}
  \newcommand{\GG}[1]{{\color{cyan} GG: #1}}
  \newcommand{\ang}[1]{{\color{magenta} Ang: #1}}
  \newcommand{\saaket}[1]{{\color{purple} Saaket: #1}}
\else
  \newcommand{\jc}[1]{}
  \newcommand{\eric}[1]{}
  \newcommand{\richard}[1]{}
  \newcommand{\vincent}[1]{}
  \newcommand{\GG}[1]{}
  \newcommand{\ang}[1]{}
  \newcommand{\saaket}[1]{}
\fi

\newcommand{\defeq}{\vcentcolon=}

\newcommand{\Ical}{\mathcal{I}}

\title{On the Reliability of Computer Use Agents}

\author{
Gonzalo Gonzalez-Pumariega,
Saaket Agashe,
Jiachen Yang,
Ang Li,
Xin Eric Wang
}

\correspondingauthor{eric@simular.ai}

\usepackage[authoryear, sort&compress, round]{natbib}
\uselogo{}
\renewcommand{\today}{}

\begin{abstract}
    Computer-use agents have rapidly improved on real-world tasks such as web navigation, desktop automation, and software interaction, in some cases surpassing human performance. Yet even when the task and model are unchanged, an agent that succeeds once may fail on a repeated execution of the same task. This raises a fundamental question: if an agent can succeed at a task once, what prevents it from doing so reliably? In this work, we study the sources of unreliability in computer-use agents through three factors: stochasticity during execution, ambiguity in task specification, and variability in agent behavior. We analyze these factors on OSWorld using repeated executions of the same task together with paired statistical tests that capture task-level changes across settings. Our analysis shows that reliability depends on both how tasks are specified and how agent behavior varies across executions. These findings suggest the need to evaluate agents under repeated execution, to allow agents to resolve task ambiguity through interaction, and to favor strategies that remain stable across runs.
\end{abstract}

\begin{document}
\maketitle

\section{Introduction}

\begin{figure}[t]
    \centering
    \includegraphics[width=\linewidth]{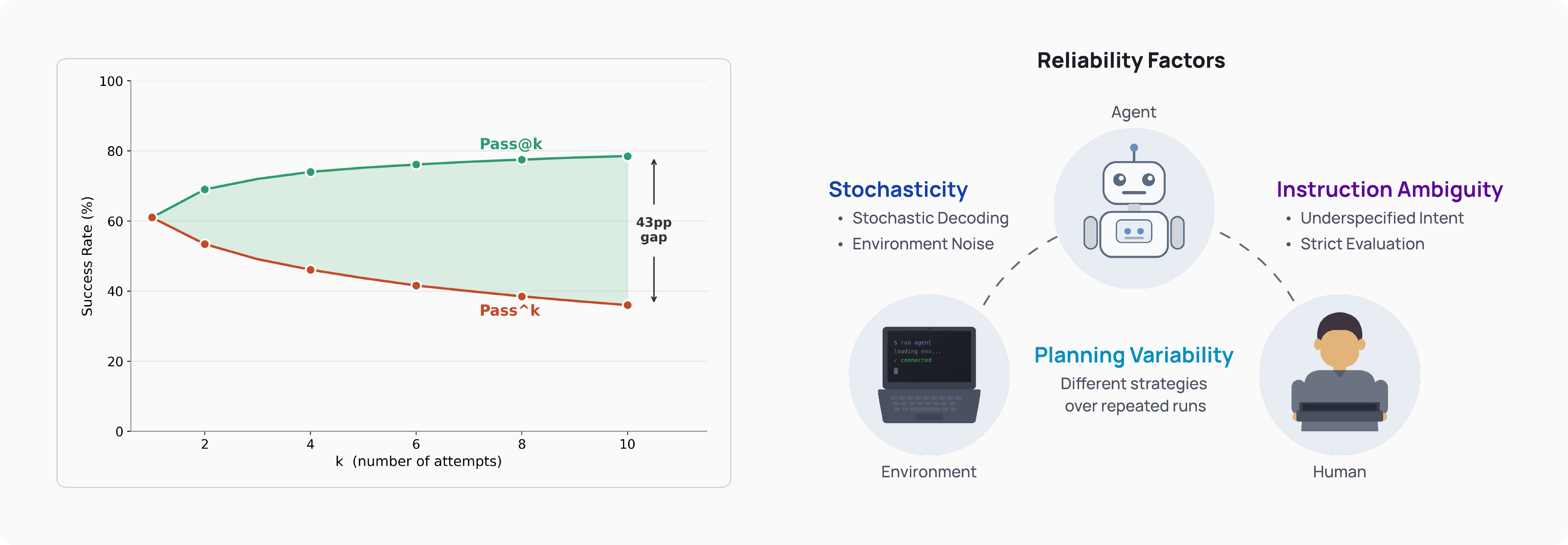}
    \caption{(left) Performance of a strong computer-use agent (Agent S3 with GPT-5) across repeated attempts. While Pass@10 reaches approximately 78\%, the corresponding \passk{10} indicates that the agent succeeds on all 10 executions for only about 36\% of tasks, indicating that achieving reliability across repeated executions is challenging. (right) Overview of factors that contribute to unreliability. We decompose task execution into three components: stochasticity during execution, ambiguity in task specification, and variability in agent behavior, which adversely affect reliability across repeated executions.}
    \label{fig:overview_teaser}
\end{figure}

Recent advances in computer-use agents have enabled strong performance on benchmark tasks. On environments such as OSWorld \citep{xie2024osworld}, modern agents can solve a higher number of tasks than human baselines \citep{claudesonnet4_6,agents3,kimik2_5}. These results suggest that agents are increasingly capable of performing tasks across diverse environments, including operating systems, web interfaces, and productivity software. However, previously reported results in single-run, averaged multi-run, or Best-of-N settings do not capture how reliably agents behave on repeated runs of each individual task (Figure~\ref{fig:overview_teaser} (left)). In practice, an agent that succeeds once may fail when the same task is executed again, revealing the problem of reliability. 

In real-world settings, reliability is a requirement for practical deployment across domains such as aerospace software \citep{do178c}, electrical and electronic systems \citep{iec61508}, automotive systems \citep{iso26262}, medical software \citep{iec62304}, and regulated AI systems \citep{Tabassi_2023,EU_HLEG_TrustworthyAI_2019}, where reproducible behavior is critical for safety and trust. In these settings, it is not sufficient for an agent to succeed at a task once within multiple attempts. Yet in practice, computer-use agents often exhibit unreliable behavior even when the task and model are unchanged \citep{agents3}, succeeding in one run but failing in another. This raises a fundamental question: \textbf{if an agent can succeed at a task once, what prevents it from doing so reliably}?

To understand where reliability fails, we decompose task execution into three key components: (1) task specification, (2) agent decision-making, and (3) stochasticity during execution (Figure~\ref{fig:overview_teaser} (right)). First, task instructions may be underspecified, admitting multiple valid interpretations that do not always align with evaluation criteria. Second, even when tasks are clearly specified, the agent may adopt different strategies for completing the task, some of which are more robust than others. Third, stochasticity in decoding or small changes in the environment can alter the trajectory of execution. Motivated by this decomposition, we study the effects of stochasticity, instruction ambiguity, and planning variability on reliability through controlled experiments on OSWorld.

To analyze the sources of unreliability, we first establish how to measure it. 
Metrics such as Pass@k \citep{chen2021codex} measure whether an agent can succeed at least once across multiple attempts, effectively rewarding the ability to produce a single successful outcome and therefore do not capture reliability. More recent work introduces repeated-run success metrics such as \passk{k} \citep{yao2024tau}, which quantify the probability that repeated executions of the same task all succeed. While this metric better aligns with reliability, it summarizes outcomes across tasks and does not capture how reliability varies at the level of individual tasks or how it changes across settings. In this work, we build on \passk{k} with a paired statistical analysis of repeated executions, enabling us to characterize task-level changes and detect both improvements and regressions in reliability.

The decomposition of task execution has several implications for the evaluation and design of computer-use agents. First, reliability cannot be inferred from the ability to solve a task once and must instead be evaluated across repeated executions. Second, instruction ambiguity highlights a key challenge, as such ambiguity is unavoidable in real-world tasks and may require agents to resolve uncertainty through interaction rather than relying on fixed task specifications. Third, the presence of multiple valid strategies indicates that variability in agent behavior is itself a source of unreliability, suggesting the need to favor strategies that remain stable across executions. Together, these observations point to the need for agents and evaluation settings that explicitly account for behavior across repeated runs, rather than focusing solely on one successful outcome.

\section{CUA Reliability Evaluation}

To study how different sources of variation affect reliability across repeated executions, we aim to answer the following research questions:

\begin{enumerate}
    \item \textbf{Stochasticity}: To what extent does stochasticity in decoding and environment dynamics contribute to unreliable outcomes across repeated executions?
    
    \item \textbf{Instruction Ambiguity}: To what extent does ambiguity in task specification, and its mismatch with evaluation criteria, contribute to reliability?

    \item \textbf{Planning Variability}: To what extent does variability in agent strategies across executions drive unreliability, beyond effects due to instruction ambiguity?
\end{enumerate}

\subsection{Problem Formulation}
To analyze reliability across repeated executions of the same task, we formalize computer-use agents (CUAs) within a POMDP framework. Specifically, we model task execution as a partially observable Markov Decision Process (POMDP) defined as $\mathcal{M} = \langle \mathcal{S}, \mathcal{O}, \mathcal{A}, \mathcal{T}, \Ical, R \rangle$, where $\mathcal{S}$ is the state space encoding the computer state, $\mathcal{O}$ is the observation space such as desktop screenshots, $\mathcal{A}$ is the action space of the agent (e.g. $\texttt{agent.click(...)}$ and $\texttt{agent.type(...)}$), $\mathcal{T}: \mathcal{S} \times \mathcal{A} \rightarrow \Delta(\mathcal{S})$ is a stochastic transition function, $\Ical$ is the space of possible user instructions represented in natural language, and $R: (\mathcal{S} \times \mathcal{A})^* \times \Ical \rightarrow [0,1]$ denotes the instruction reward function that assigns a scalar reward to a trajectory $\tau \defeq (s_0,a_0,\dotsc,a_{T-1},s_T)$ conditioned on instruction $I \in \Ical$. A task is defined as $x \defeq (s_0, I) \in \mathcal{X}$, consisting of an initial state and instruction.

Standard approaches optimize for expected success, maximizing $\mathbb{E}_{\tau \sim \pi(\cdot \mid x)}[R(\tau, x)]$, where $R(\tau, x) \in [0,1]$ denotes a scalar reward for a single execution of task $x$. For simplicity in evaluating reliability, we consider a binary success outcome $r_{x,j} = R(\tau_j,x)$ such that $r_{x,1}, \dotsc, r_{x,n} \in \{0,1\}$ denote the outcomes of $n$ executions of the policy on the same task $x$.

Under this formulation, success corresponds to the probability that a single execution succeeds, $\mathbb{E}_{\tau \sim \pi(\cdot \mid x)}[r_{x,1}] = \Pr(r_{x,1} = 1)$. In contrast, we define reliability as the probability that all executions of the same task succeed, given by $\mathbb{E}_{\tau_1, \dotsc, \tau_n \sim \pi(\cdot \mid x)}\!\left[\prod_{j=1}^n r_{x,j}\right] = \Pr(r_{x,1} = 1, r_{x,2} =1,\cdots , r_{x,n} = 1)$.

\subsection{Reliability Metrics}

\begin{figure}[t]
    \centering
    \includegraphics[width=\linewidth]{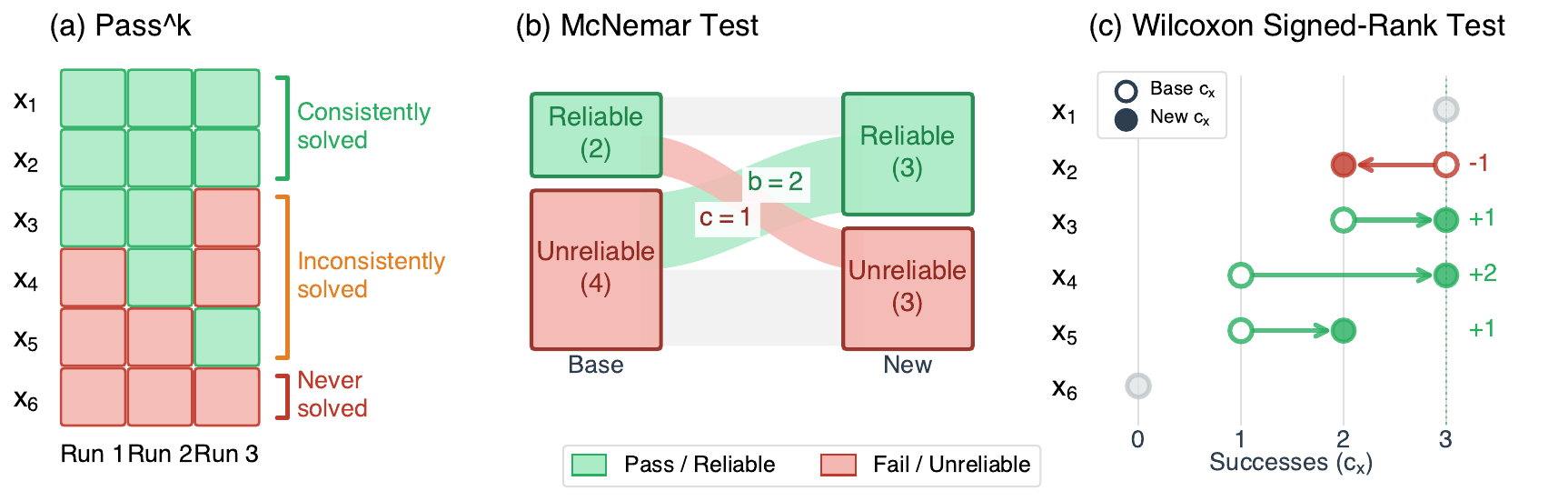}
    \caption{We illustrate three metrics for analyzing consistency in agent performance over multiple runs of the same task. (a) \passk{k} (repeated-run success) estimates the probability that $k$ executions of a task succeed, averaged across tasks. (b) McNemar measures improvements and regressions in reliability between two settings by counting tasks that transition between being consistently solved and not. (c) Wilcoxon signed-rank test compares per-task success counts across settings, capturing incremental changes in consistency even when full reliability is not achieved.}
    \label{fig:intuition}
\end{figure}

Given the definition of reliability as reproducible success across repeated executions, we seek metrics that characterize how consistently an agent succeeds on the same task. For each task $x$, we execute the policy $n$ times, yielding binary outcomes $r_{x,1}, \dots, r_{x,n} \in \{0,1\}$ and $c_x = \sum_{j=1}^n r_{x,j}$ successful runs. These repeated outcomes allow us to categorize tasks as \emph{consistently solved} ($c_x = n$), \emph{inconsistently solved} ($0 < c_x < n$), or \emph{never solved} ($c_x = 0$), providing a structured view of variability across runs.

\paragraph{\passk{k} (Repeated-Run Success).}
Following \cite{yao2024tau}, we adopt the metric
\[
\text{\passk{k}} = \mathbb{E}_{x \sim \mathcal{X}} \left[ \frac{\binom{c_x}{k}}{\binom{n}{k}} \right],
\]
which estimates the probability that $k$ executions of a task all succeed. We highlight two cases: (1) \passk{1}, the marginal success rate across executions, given by $\text{\passk{1}} = \mathbb{E}_{x \sim \mathcal{X}} \left[ \frac{c_x}{n} \right]$, which captures agent capability and (2) \passk{n}, the fraction of tasks that succeed on all repeated executions, given by $\text{\passk{n}} = \mathbb{E}_{x \sim \mathcal{X}} \left[ \mathbf{1}[c_x = n] \right]$, which directly measures reliability as reproducible success.

While \passk{k} summarizes performance across tasks, it does not capture how outcomes change on the same tasks across different settings. To study how reliability varies across settings and identify improvements or regressions at the task level, we use paired statistical tests.

\paragraph{McNemar Test (Reliability Transitions).}
We define a binary indicator $z_x = \mathbf{1}[c_x = n]$ for whether task $x$ is consistently solved. To compare two settings, we apply McNemar’s test~\citep{mcnemar} to paired outcomes $\{z_x^{(\text{base})}, z_x^{(\text{new})}\}$. Let
\[
b = \sum_x \mathbf{1}[z_x^{(\text{base})} = 0,\ z_x^{(\text{new})} = 1], \quad
c = \sum_x \mathbf{1}[z_x^{(\text{base})} = 1,\ z_x^{(\text{new})} = 0]
\]
denote the number of tasks that improve and regress, respectively. The test statistic is
\[
\chi^2 = \frac{(b - c)^2}{b + c},
\]
which measures the imbalance between improvements and regressions. This test evaluates whether more tasks are reliably solved than not. We report $(b-c)$, which indicates the direction of change, and compute p-values using the corresponding $\chi^2$ statistic, with significance assessed at $p < 0.05$.

\paragraph{Wilcoxon Signed-Rank Test (Consistency Improvements).}
To capture partial improvements in repeated-run performance, we compare per-task success counts across settings. For each task, we compute differences $d_x = c_x^{(\text{new})} - c_x^{(\text{base})}$ and apply the Wilcoxon signed-rank test~\citep{wilcoxon}. Let $\{d_x\}$ denote the set of nonzero differences, and let $R_x$ be the rank of $|d_x|$ among these values (with ties assigned average ranks). The test statistic is
\[
W = \sum_{d_x > 0} R_x,
\]
which measures whether improvements tend to be larger or more frequent than regressions. This detects incremental improvements in consistency (e.g., $1 \to 2$ or $2 \to 3$ successes), even when full reproducibility is not achieved. We report $\Delta c_x = \frac{1}{|\mathcal{X}|} \sum_{x \in \mathcal{X}} d_x$, the average change in per-task success counts, and compute p-values from the Wilcoxon signed-rank test using the test statistic $W$, with significance assessed at $p < 0.05$.

\subsection{Models}

We evaluate reliability across a range of computer-use agents spanning both frontier and open-source models. We primarily consider strong frontier models, including GPT-5 \citep{gpt5}, Claude Sonnet 4.6 \citep{claudesonnet4_6}, and Kimi 2.5 \citep{kimik2_5}, to assess whether reliability challenges persist in high-performing systems. We additionally include smaller and open-source models, such as Qwen-3VL-8B-Instruct \citep{qwen3vl}, OpenCUA \citep{opencua}, and UI-TARS-1.5-7B \citep{uitars}, to enable controlled experiments and examine how reliability varies with model capability. For models with provided execution setups, we use the default running scripts in OSWorld \citep{xie2024osworld}. Otherwise, we adopt the Agent S3 \citep{agents3} settings for action space and grounding and add an (S3) suffix. All models use API required temperature (e.g. temperature 1 for GPT-5) or temperature 0.7 unless otherwise specified.

\section{Stochastic Decoding and Execution Noise}

We begin by investigating whether reliability is affected by stochasticity in decoding and execution. A natural hypothesis is that variability across runs is primarily driven by randomness in token sampling or environment dynamics, and that enforcing determinism should therefore improve reliability. To test this, we introduce interventions that probe stochasticity at two levels: (1) agent-side determinism, which removes variability in both token sampling and strategy by using temperature-0 decoding with batch-invariant inference and constraining the agent to follow a fixed high-level plan across runs, and (2) controlled environment perturbations, which introduce non-functional variations in observations to evaluate how reliability changes under these variations. By evaluating repeated executions of the same tasks under these settings, we isolate the extent to which stochasticity alone accounts for differences in reliability.

\subsection{Deterministic Agent Execution}
\label{sec:deterministic}

\begin{table}[h]
\centering
\small
\setlength{\tabcolsep}{5pt}
\begin{tabular}{llcccc}
\toprule
\textbf{Model} & \textbf{Setting} & \textbf{\passk{1}} & \textbf{\passk{3}} & \boldmath$b-c$ & \boldmath$\Delta c_x$ \\
\midrule

Qwen (S3) & Baseline      & 0.329 & 0.222 & --   & --       \\
Qwen (S3) & Deterministic & 0.293 & 0.166 & -20* & -0.108*  \\
Qwen (S3) & Strategy Determinism & 0.313 & 0.219 & -1   & -0.047   \\

\midrule

OpenCUA & Baseline      & 0.226 & 0.125 & --   & --      \\
OpenCUA & Deterministic & 0.259 & 0.180 & 20*  & 0.097*  \\
OpenCUA & Strategy Determinism & 0.262 & 0.188 & 23*  & 0.108*  \\

\midrule

UI-TARS-1.5 & Baseline      & 0.253  & 0.152 & --   & --      \\
UI-TARS-1.5 & Deterministic & 0.290 & 0.205 & 19*  & 0.111*  \\
UI-TARS-1.5 & Strategy Determinism & 0.264 & 0.169 & 14* & 0 \\

\bottomrule
\end{tabular}
\caption{Reliability metrics across deterministic decoding strategies. McNemar and Wilcoxon statistics are computed relative to the stochastic Baseline setting. An asterisk (*) denotes statistical significance at $p < 0.05$ under either test.}
\label{tab:determinism_summary}
\end{table}

We first evaluate whether removing sampling randomness improves reliability by comparing a stochastic decoding baseline to deterministic decoding with temperature 0 and batch-invariant inference. For Qwen, it leads to statistically significant regressions, with many tasks transitioning from reliable to not ($b-c=-20$); however, there are significant improvements for OpenCUA ($b-c=20$) and UI-TARS-1.5 ($b-c=19$). This contrast suggests that eliminating token-level stochasticity does not always improve reliability and is dependent on model-specific decoding behavior. Regardless, all models experience drops from \passk{1} to \passk{3} which indicates that removing sampling randomness alone is insufficient to ensure fully reliable execution.

We next evaluate whether enforcing determinism at the level of high level strategy, in addition to deterministic decoding, improves reliability. Instead of allowing the agent to replan on each execution, the agent is separately prompted to generate a plan (Appendix~\ref{prompt:sd:sample_plan}) that is reused in future repeated runs (Appendix~\ref{prompt:sd:fix_plan}). For OpenCUA and UI-TARS-1.5, it maintains significant improvements in reliability transitions achieved under deterministic decoding, while for Qwen it mitigates the regression introduced by deterministic decoding, yielding more balanced transitions $(b-c=-1)$. However, relative to the stochastic baseline, we observe little to no improvement for Qwen and UI-TARS-1.5 and only minor gains for OpenCUA. These results suggest that constraining the agent’s high level strategy can mitigate some of the instability introduced by deterministic decoding, but does not consistently improve reliability relative to the stochastic baseline. Thus, sampling and conditioning on a fixed plan to stabilize deterministic execution is insufficient to improve reliability.

\subsection{Sensitivity to Environment Noise}

\begin{table}[h]
\centering
\small
\setlength{\tabcolsep}{5pt}
\begin{tabular}{llcccc}
\toprule
\textbf{Model} & \textbf{Condition} & \textbf{\passk{1}} & \textbf{\passk{3}} & \boldmath$b-c$ & \boldmath$\Delta c_x$ \\
\midrule

GPT-5 (S3) & Baseline & 0.576 & 0.454 & -- & -- \\
GPT-5 (S3) & Perturbed & 0.561 & 0.416 & -10 & -0.033 \\

\midrule

Claude & Baseline & 0.702 & 0.612 & -- & -- \\
Claude & Perturbed & 0.675 & 0.557 & -20* & -0.080* \\

\midrule

Kimi & Baseline & 0.508 & 0.357 & -- & -- \\
Kimi & Perturbed & 0.516 & 0.355 & -1 & 0.025 \\

\bottomrule
\end{tabular}
\caption{Reliability metrics under environment perturbation. McNemar and Wilcoxon statistics are computed relative to the fixed environment Baseline setting. An asterisk (*) denotes statistical significance at $p < 0.05$ under either test.}
\label{tab:environment_stochasticity}
\end{table}

We evaluate whether agents are sensitive to environment noise, particularly non-functional task differences in environmental observations. We compare against (1) a baseline consisting of three repeated runs in an unperturbed environment and (2) a cross-environment setting where the first baseline run is paired with runs in two perturbed environments with cosmetic differences (details in Appendix~\ref{app:perturbation_details}). Table~\ref{tab:environment_stochasticity} summarizes the results; GPT-5 regresses in reliability transitions ($b-c=-10$) while Claude has a significant drop ($b-c=-20$). In the case of Kimi, we observe that \passk{3} is already low even within the same environment, so further perturbations have minimal effects.

\section{Instruction Ambiguity}

We next investigate whether reliability is affected by ambiguity in task instructions. When instructions are underspecified, they may admit multiple valid interpretations, while evaluators often expect a more specific outcome. As a result, an agent may follow different reasonable strategies across runs, only some of which satisfy the evaluation criteria. We test whether resolving this ambiguity improves reliability through two interventions: (1) clarifying task instructions before execution to make success criteria more explicit, and (2) providing feedback during execution using an LLM-based user simulator that identifies mismatches between the agent’s behavior and the expected outcome. These interventions allow us to evaluate how reducing ambiguity at different stages of execution affects reliability.

\subsection{Clarification Before Execution}
\label{sec:clarification}

\begin{figure}[h]
    \centering
    \includegraphics[width=\textwidth]{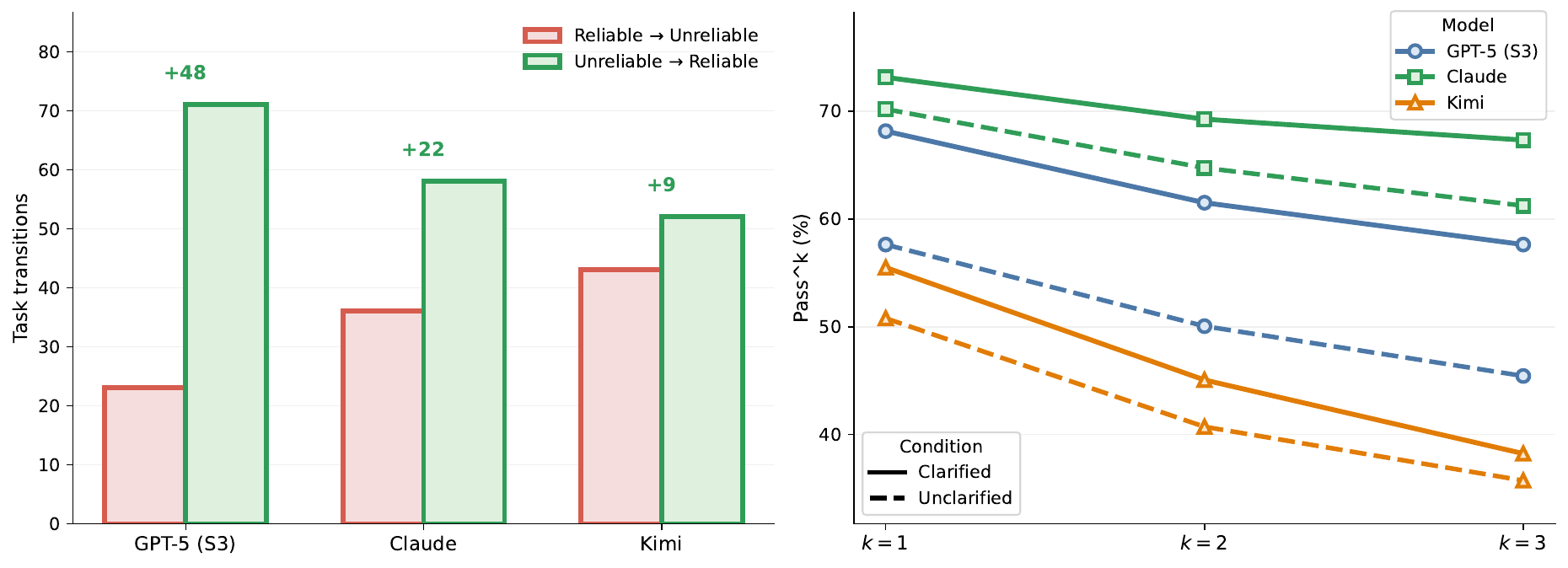}
    \caption{Task-level transitions in reliability under instruction clarification measured using McNemar analysis (left), and repeated-run success under clarified and unclarified instructions measured using \passk{k} (right).}
    \label{fig:clarification}
\end{figure}

We evaluate whether resolving instruction ambiguity improves reliability by comparing performance on unclarified and clarified task descriptions. To construct clarified instructions, we rewrite task descriptions to make success criteria explicit using information from the evaluation script and associated functions, while avoiding additional details that would trivialize the task. 
Details of the prompting and minimal human corrections are provided in Appendix~\ref{app:clarification_details}. 
We find that clarification leads to consistent improvements across models, with more tasks transitioning from not reliably solved to reliably solved than the reverse (Figure~\ref{fig:clarification}). These gains are observed across GPT-5, Claude, and Kimi, along with increases in both \passk{1} and \passk{3}. For Kimi, improvements in reliability are smaller, but per-task success counts increase significantly (Table~\ref{tab:instruction_ambiguity},$\Delta c_x=0.141)$, indicating gradual progress toward reliable execution. Overall, these results show that ambiguity in task instructions contributes to reduced reliability, and that making success criteria explicit improves performance across repeated executions.

\subsection{Clarification During User Interaction}
\label{sec:interaction}

\begin{table}[h]
\centering
\small
\setlength{\tabcolsep}{5pt}
\begin{tabular}{llcccc}
\toprule
\textbf{Model} & \textbf{Setting} & \textbf{\passk{1}} & \textbf{\passk{3}} & \boldmath$b-c$ & \boldmath$\Delta c_x$ \\
\midrule

GPT-5 (S3) & Original         & 0.576 & 0.454  & --  & --      \\
GPT-5 (S3) & Clarified        & 0.681 & 0.576  & 48* & 0.327*  \\
GPT-5 (S3) & Retry (Binary)   & 0.645 & 0.557  & 41* & 0.216*  \\
GPT-5 (S3) & Retry (Clarify)  & 0.730 & 0.632  & 68* & 0.474*  \\

\midrule

Claude & Original             & 0.702 & 0.612 &  --  & --     \\
Claude & Clarified            & 0.731 & 0.673 &  22* & 0.089  \\
Claude & Retry (Binary)       & 0.732 & 0.668 &  20* & 0.091  \\
Claude & Retry (Clarify)      & 0.774 & 0.706 &  34* & 0.216* \\

\midrule

Kimi & Original               & 0.508 & 0.357 & --   & --     \\
Kimi & Clarified              & 0.555 & 0.382 & 9    & 0.141* \\
Kimi & Retry (Binary)         & 0.623 & 0.496 & 50*  & 0.346* \\
Kimi & Retry (Clarify)        & 0.727 & 0.634 & 100* & 0.657* \\

\bottomrule
\end{tabular}
\caption{Reliability metrics across instruction ambiguity interventions. McNemar and Wilcoxon statistics are computed relative to the Original baseline setting. An asterisk (*) denotes statistical significance at $p < 0.05$ under either test.}
\label{tab:instruction_ambiguity}
\end{table}

Clarification before execution may miss ambiguities that only become apparent during execution for a given agent and model. To address this, we introduce a user simulator that provides targeted feedback on failed executions based on the agent’s trajectory, task instruction, and evaluation signals 
(details in Appendix~\ref{app:user_simulator_prompts})
. We evaluate this setting, Retry (Clarify), against a retry baseline without targeted feedback, Retry (Binary), to isolate the effect of feedback content from additional attempts. Both retry baselines are allowed up to 5 retries. Retry (Clarify) consistently outperforms Retry (Binary) across all metrics and models (Table~\ref{tab:instruction_ambiguity}). We also observe that a single execution with clarified instructions from Section~\ref{sec:clarification} can match or exceed the performance of Retry (Binary), indicating that resolving ambiguity can be more effective than repeated attempts alone. These results show that clarifying ambiguity during execution through targeted feedback is an effective mechanism for improving reliability.

\section{Planning Variability}

While clarifying task instructions improves performance and reliability, agents may still exhibit inconsistent behavior across repeated executions of the same task, even when the task is well-specified. This suggests that unreliability may arise from variability in the strategies selected by the agent, where different executions follow different plans with varying robustness. To test this hypothesis, we design controlled interventions that incorporate information from prior executions to guide subsequent runs, allowing us to evaluate whether stabilizing or refining strategies improves consistency. Starting from an initial setting (Iteration 0) with clarified instructions (Section~\ref{sec:clarification}), we incorporate information from prior rollouts to guide the next execution (Iteration 1), and further refine this guidance using additional rollouts in subsequent iterations (Iteration 2). This setup allows us to evaluate both the immediate impact of incorporating prior experience and how iteratively refining strategies reduces planning variability and improves reliability. 

\begin{table}[t]
\centering
\small
\setlength{\tabcolsep}{5pt}
\begin{tabular}{llcccc}
\toprule
\textbf{Model} & \textbf{Iteration} & \textbf{\passk{1}} & \textbf{\passk{3}} & \boldmath$b-c$ & \boldmath$\Delta c_x$ \\
\midrule

GPT-5 (S3) & Iteration 0 & 0.681  & 0.576 & -- & -- \\
GPT-5 (S3) & Iteration 1 & 0.710 & 0.618  & 15 & 0.086*   \\
GPT-5 (S3) & Iteration 2 & 0.725 & 0.651 & 27* & 0.130*   \\

\midrule

Claude & Iteration 0 & 0.731 & 0.673 & -- & -- \\
Claude & Iteration 1 & 0.713 & 0.632 & -15* & -0.055* \\
Claude & Iteration 2 & 0.730 & 0.662 & -4 & -0.006 \\

\midrule

Kimi & Iteration 0 & 0.555 & 0.382 & -- & -- \\
Kimi & Iteration 1 & 0.567 & 0.435 & 19* & 0.036 \\
Kimi & Iteration 2 & 0.563 & 0.424 & 15 & 0.025 \\

\bottomrule
\end{tabular}
\caption{Reliability metrics across iterations of plan extraction and iterative plan refinement. McNemar and Wilcoxon statistics are computed relative to Iteration 0. An asterisk (*) denotes statistical significance at $p < 0.05$ under either test.}
\label{tab:planning_variability}
\end{table}

\subsection{Effect of Plan Extraction}
\label{sec:plan_extraction}
We first analyze the effect of incorporating information from prior executions through plan extraction. Starting from an initial execution setting (Iteration 0), we perform multiple rollouts of the same task and extract structured feedback from these trajectories, including both successful behaviors and recurring failure patterns (Appendix~\ref{app:pe_pr:base_prompt}), to synthesize a plan that guides the next execution (Iteration 1). We modify the Behavior Judge \cite{agents3} to generate feedback over behavior narrative representations of  trajectories.
In cases where all rollouts are successful, we do not provide feedback since the model is already reliable on the task. If all rollouts fail, we generate feedback based on partial success to encourage exploration beyond previously unsuccessful strategies (Appendix~\ref{app:pe_pr:extract_all_failures}). We assume access to ground-truth signals to label rollout task success to avoid confounds from imperfect judge signals. Unlike strategy determinism (Section~\ref{sec:deterministic}), which samples a plan given the task instruction and initial screenshot, plan extraction derives a plan from prior rollouts to reuse in future repeated runs (Appendix~\ref{app:pe_pr:plan_feedback_addon}).

We summarize results from Iteration 0 to Iteration 1 in Table~\ref{tab:planning_variability}. We find that incorporating feedback from prior executions improves \passk{1} and \passk{3} for both GPT-5 (by $2.9\%\text{ and }4.2\%$) and Kimi (by $1.2\%\text{ and }5.3\%$). For GPT-5, we find statistically significant gains under the Wilcoxon signed-rank test ($\Delta c_x = 0.086$), indicating that partial improvements towards reliability were achieved, while Kimi achieves significant gains under McNemar’s test ($b-c=19$), with more tasks becoming reliably solved than not. In contrast, Claude exhibits significant regressions when incorporating prior feedback $(b-c=-15,\Delta c_x=-0.055)$, suggesting that extracted plans can introduce instability when they are not well aligned with reliable strategies. 
We hypothesize this is because Claude is biased towards code solutions which leads to unseen environment changes that Behavior Judge struggles to detect, producing incomplete feedback. Overall, these results indicate that variability in agent strategies is a key driver of unreliability, and that guiding execution with feedback from prior rollouts can help mitigate it.

\subsection{Effect of Iterative Plan Refinement}
\label{sec:plan_refinement}

We next analyze the effect of iteratively refining plans across multiple rounds of execution. Following Iteration 1, where executions follow a plan extracted from prior rollouts (Section~\ref{sec:plan_extraction}), the agent is run again multiple times and the new feedback is used to update the existing plan using the initial plan extraction prompt (Appendix~\ref{app:pe_pr:refinement_addon}), producing a refined plan for Iteration 2. In cases where all rollouts failed, successful rollouts from previous iterations can be utilized for plan extraction (Appendix~\ref{app:pe_pr:extract_historical_success}).

We summarize overall results from Iteration 0 to Iteration 2 in Table~\ref{tab:planning_variability}. GPT-5 continues to improve across iterations, with gains remaining statistically significant under both McNemar’s and Wilcoxon signed-rank tests $(b-c=27,\Delta c_x=0.130)$. Kimi also shows overall improvements in \passk{1} and \passk{3} ($0.8\%\text{ and }4.2\%$), though it exhibits a slight regression relative to Iteration 1. In contrast, Claude continues to underperform relative to its initial setting, although the regressions in McNemar and Wilcoxon are no longer statistically significant $(b-c=-4,\Delta c_x=0.006)$, suggesting that the reliability loss was stabilized through the refined plans. Overall, these results suggest that iterative refinement can further improve reliability when feedback is effectively incorporated, though stronger improvements across models may require additional iterations for plans to better reflect reliable execution patterns.

\section{Discussion and Conclusion}
We summarize the key findings from our experiments and their implications for improving reliability in computer-use agents.

\paragraph{Sensitivity to Stochasticity}
We find that (1) enforcing deterministic decoding does not consistently improve reliability across models, (2) constraining agents to follow fixed strategies stabilizes execution but does not resolve reliability issues, and (3) introducing non-functional environment perturbations degrades reliability despite no change in task correctness. These results suggest that removing stochasticity limits the agent’s ability to adapt to small variations in execution, making failures more likely under even minor changes in the environment. While fully deterministic approaches such as symbolic programs can achieve high reliability under fixed conditions, they remain brittle to environmental variation and can fail catastrophically. A promising direction is to combine symbolic structure with stochastic decoding agents, using symbolic representations to guide execution while allowing stochasticity to enable adaptation under environmental variation \citep{gao2023palprogramaidedlanguagemodels,wang2024executablecodeactionselicit,chen2023programthoughtspromptingdisentangling}

\paragraph{Instruction Ambiguity} We find that (1) clarifying task instructions before execution leads to substantial improvements in reliability, and (2) incorporating targeted feedback during execution is more effective than both retry-based baselines and static clarification. These results suggest that reliable behavior cannot be achieved through static instruction specification alone, and instead requires agents to actively resolve ambiguity during execution. This can be supported by enabling benchmarks to incorporate user simulators that provide targeted clarification when needed, rather than relying solely on fixed task descriptions. This framing also makes approaches such as active preference elicitation increasingly relevant for improving reliability \citep{wang2024apricotactivepreferencelearning,handa2024bayesianpreferenceelicitationlanguage,piriyakulkij2024activepreferenceinferenceusing}

\paragraph{Planning Variability} We find that (1) enforcing fixed strategies across runs stabilizes execution but does not consistently improve reliability over stochastic baselines, and (2) incorporating information from prior executions through plan extraction and iterative refinement can improve reliability, though the gains vary across models and depend on the quality of the guidance. These results suggest that variability in planning remains a key challenge, and that improving reliability requires methods that can consistently identify and follow reliable execution strategies across runs. Prior work has explored related mechanisms in isolation, including iterative refinement from prior executions \citep{shinn2023reflexionlanguageagentsverbal,madaan2023selfrefineiterativerefinementselffeedback} and leveraging multiple prior trajectories as in-context demonstrations to guide behavior \citep{gupta2025leveragingincontextlearninglanguage}. A promising direction is to develop approaches that better combine these ideas by leveraging information across multiple trajectories to guide future behavior while remaining adaptable to new contexts.

Taken together, our results show that achieving reliable behavior in computer-use agents requires addressing multiple interacting sources of variation, including stochasticity, instruction ambiguity, and planning variability. As these agents are increasingly deployed in real-world settings, reliability becomes a critical requirement rather than an auxiliary metric, since inconsistent behavior can undermine both usability and trust. Our findings highlight the importance of evaluating agents under repeated execution and incorporating mechanisms such as interaction and guidance from prior executions when studying reliability. We hope this work motivates future research on building computer-use agents that are not only capable of solving tasks, but do so reliably across repeated runs.

\newpage
\bibliography{main}

\appendix
\section{Appendix}
\subsection{Use of LLMs}

We used GPT-5 to assist with writing by generating structured sentences from ideas and improving writing flow. We used a combination of GPT-5 and Claude to generate code for figures and LaTeX tables (while inputting data manually).

\section{Related Work}

\paragraph{Computer-Use Agents and Benchmarks.}
A central goal of recent work is to build agents that can execute real-world tasks by interacting directly with computing environments. OSWorld provides a comprehensive setting for general computer use, requiring agents to operate across applications, file systems, and web environments \citep{xie2024osworld}, while a growing body of work studies related settings such as robust and long-horizon GUI interaction \citep{zhao2026worldgui,wu2026osmarathon}, web navigation \citep{zhou2024webarena,koh2024visualwebarena,deng2023mind2web}, and enterprise software \citep{ai2025prosoftarena,dai2025scuba}. Despite differences in environment and task design, these benchmarks primarily report single-run performance metrics such as success rate or task completion, rather than measuring consistency across repeated executions. In contrast, we focus on reliability as a primary evaluation target, measuring whether agents can consistently succeed on the same task across multiple runs.

\paragraph{Measuring Reliability in Agents.}

Evaluation of agents is commonly based on sampling-based metrics that estimate success across multiple attempts, with Pass@k being one of the most widely used metrics for large language models. Pass@k measures whether an agent can produce at least one successful outcome across $k$ independent samples, and is often interpreted as an upper bound on task success under repeated sampling \citep{chen2021codex}. However, this metric does not capture whether an agent can succeed consistently, as it only requires success in a single attempt. More recently, $\tau$-bench \citep{yao2024tau} introduces \passk{k} as a reliability metric, defined as the probability that an agent succeeds across all $k$ repeated executions of the same task, which can be viewed as a lower bound on consistent performance. While \passk{k} captures consistency, it remains an aggregate metric and does not capture how reliability changes at the level of individual tasks across settings. In this work, we build upon this metric with paired statistical tests over repeated executions, allowing us to detect per-task improvements and regressions in consistency and to compare reliability changes across conditions.

\section{System Prompts}
\label{app:system-prompt}

\subsection{Strategy Determinism}

\subsubsection{Sampling Plan Prompt}
\begin{lstlisting}[style=prompt,caption={Strategy Determinism Plan Sampling},label={lst:sys-prompt},numbers=none]
You are an expert at creating step-by-step plans for GUI automation tasks.

Task: {instruction}

Please analyze the current screenshot and create a detailed step-by-step plan to accomplish this task.

Your plan should:
1. Be specific about what to click, type, or interact with
2. Include the sequence of actions needed  
3. Be detailed enough that another agent could follow it

Respond with a numbered list of steps:
1. [Action description]
2. [Action description]
...

Only provide the plan - do not execute any actions.
\end{lstlisting}
\label{prompt:sd:sample_plan}

\subsubsection{Reusing Plan Prompt}
\begin{lstlisting}[style=prompt,caption={Strategy Determinism Plan Addon},label={lst:sys-prompt},numbers=none]
[PLAN TO FOLLOW] PRE-GENERATED STEP-BY-STEP PLAN: Below is a detailed plan for completing the task: `{instruction}`
You must follow this plan step-by-step to succeed at the task.

{plan_text}

IMPORTANT: Follow this plan closely. Each step should guide your next action. Adapt the plan to the current screen state when necessary, but stay true to the overall plan structure.
\end{lstlisting}
\label{prompt:sd:fix_plan}

\subsection{Clarification Before Execution}

\subsubsection{Instruction Clarification Prompt}
\begin{lstlisting}[style=prompt,caption={Instruction Clarification Prompt},label={lst:sys-prompt},numbers=none]
You are improving the clarity of an OSWorld evaluation task instruction. The current instruction is vague and doesn't provide enough detail for an AI agent to know exactly what to do based on the evaluation criteria.

The goal is to make MINIMAL clarifications while preserving the natural, human-like tone. Do NOT rewrite as step-by-step instructions. Only clarify ambiguous details that are checked by the evaluator but missing from the instruction.

## Current Task Configuration:
```json
{example_config}
```

## Evaluator Function Implementation(s):
```python
{func_implementations}
```

## Task:

Analyze the instruction: {instruction}

Identify what the evaluator checks that isn't clear from the instruction, then provide a minimally clarified version that:
1. **Keeps the natural human tone** - should sound like a human said it
2. **Only adds missing details** that the evaluator checks for
3. **Specifies exact file names/formats/locations** if the evaluator expects them
4. **Does NOT become a step-by-step procedure** - stay conversational

Additional criteria:
1. The clarified instruction will replace the original instruction - make sure to retain essential details to avoid losing current context (e.g. do not omit links).
2. For tasks that have exact file checks, be very careful in accidentally adding or preventing information to the instruction that would cause it to no longer match the evaluator's criteria.

Format your response as:
<thoughts>
[Think through what's ambiguous and what the evaluator actually checks for]
</thoughts>

<answer>
[The minimally clarified instruction in natural language]
</answer>
\end{lstlisting}
\label{prompt:ic:clarification}

\subsubsection{Clarification Failure Analysis Prompt}

\begin{lstlisting}[style=prompt,caption={Clarification Failure Analysis Prompt},label={lst:sys-prompt},numbers=none]
You are analyzing how instruction clarifications affect agent success rate on GUI automation tasks.

You will be given:
1. ORIGINAL (before) instruction and 3 trajectory runs
2. CLARIFIED (after) instruction and 3 trajectory runs  
3. Success rate scores for all runs
4. **EVALUATOR FUNCTION**: The exact code that determines task success/failure

Your goal is to determine:
- How agent behavior changed between the original and clarified instructions
- Whether the instruction clarification had a meaningful impact on success rate
- What specific aspect of the clarification caused behavior changes

Each trajectory contains:
- **Visual Changes**: Fact captions describing GUI state changes at each step
- **Agent Reasoning**: The agent's internal thoughts and decision-making process

CRITICAL EVALUATION CONTEXT:

**Understanding the Evaluator**: The evaluator function shows EXACTLY what determines success/failure. Pay close attention to:
- **Exact matching requirements**: File names, paths, content, button states
- **Precise conditions**: Specific UI elements, file formats, directory structures  
- **Failure points**: What tiny details can break the evaluation

Many evaluators use EXACT MATCHING, where small differences cause complete failure:
- Wrong button clicked vs right button
- File saved in wrong directory
- Missing file or incorrect file name
- Different file format or content
- UI element not in expected state

ANALYSIS APPROACH:

1. **Understanding Original Success Rate**
   - Analyze the 3 'before' runs: what did successful runs do RIGHT according to the evaluator?
   - For failures: what specific evaluator requirement did they miss?

2. **Understanding Clarified Success Rate** 
   - Analyze the 3 'after' runs: what changed in agent behavior?
   - Do the new behaviors align with or conflict with evaluator requirements?

3. **Instruction Impact on Evaluator Alignment**
   - Did the clarification help agents meet evaluator requirements more precisely?
   - Did the clarification inadvertently steer agents away from evaluator requirements?
   - Are success rate changes due to better/worse evaluator alignment or just variance?

4. **Flag Assignment** 
   Choose ONE flag that best describes this case:

   **Instructions Need Manual Correction:**
   - IMPOSSIBLE_TASK: Clarification made the task impossible to complete
   - TOO_TRIVIAL: Clarification gave away the solution, making task too easy
   - HARMFUL_CONSTRAINTS: Added unnecessary constraints that hurt success rate  
   - REMOVED_HELPFUL_AMBIGUITY: Removed ambiguity the agent was successfully leveraging
   - OTHER_MISINTERPRETATION: Other instruction problem (rarely used)

   **Clarification Effect Analysis:**
   - GENUINE_IMPROVEMENT: Instruction change demonstrably helped agents meet evaluator requirements more consistently
   - RANDOM_VARIANCE: Success rate change appears unrelated to instruction, likely variance

   **Important Notes:**
   - Focus on evaluator alignment: did the clarification help or hurt the agent's ability to satisfy the exact evaluator requirements?
   - Small execution differences can completely break exact matching evaluators

   When choosing a flag, consider the success rates and evaluator requirements:
   - It does not make sense to assign GENUINE_IMPROVEMENT if the success rate did not improve
   - It does not make sense to assign IMPOSSIBLE_TASK if the success rate improved
   - If success rate decreased significantly, look for what evaluator requirement the clarification broke
   - If success rate improved, verify the clarification actually helped meet evaluator criteria (not just luck)
   
OUTPUT FORMAT:

<thinking>
[First understand the evaluator requirements thoroughly. Then analyze before/after trajectories systematically to see how agent behaviors changed relative to these requirements. Determine if success rate differences are due to better/worse evaluator alignment or variance.]
</thinking>

<answer>
Flag: [ONE of the flags above]
Analysis: [Concise explanation focusing on how the instruction change affected the agent's ability to meet the specific evaluator requirements. Explain what evaluator criteria were better/worse satisfied and why this led to the success rate change.]
</answer>
\end{lstlisting}
\label{prompt:ic:trajectory_analysis}

\subsubsection{Input for Analysis Prompt}

\begin{lstlisting}[style=prompt,caption={Clarification Failure Analysis Input},label={lst:sys-prompt},numbers=none]
Task ID: {task_id}

ORIGINAL INSTRUCTION:
{before_instruction}

CLARIFIED INSTRUCTION:
{after_instruction}

EVALUATOR FUNCTION:
```python
{evaluator_implementation}

SUCCESS RATE COMPARISON:
ORIGINAL: {before_successes}/{before_total} successes ({before_successes/before_total:.1%})
CLARIFIED: {after_successes}/{after_total} successes ({after_successes/after_total:.1%})

=== ORIGINAL INSTRUCTION TRAJECTORIES ===

--- Run {run_index} ({SUCCESS|FAILURE}, Score: {score}) ---
Visual Changes:
{fact_captions...}

Agent Reasoning:
{trajectory_reasoning...}

[repeated for each of 3 'before' runs]

=== CLARIFIED INSTRUCTION TRAJECTORIES ===

[same format for each of 3 'after' runs]

Please analyze how the instruction clarification affected agent behavior and assign the appropriate flag.
\end{lstlisting}
\label{prompt:ic:trajectory_analysis_input}

\subsection{User Simulator}
\label{app:user_simulator_prompts}

\subsubsection{User Simulator: Failure Feedback Prompt}
\begin{lstlisting}[style=prompt,caption={User Simulator Failure Feedback Prompt (Clarify Upon Retry)},label={lst:usersim-feedback},numbers=none]
You are the user who issued the following task to a computer-use agent:

"{instruction}"

The agent attempted the task but FAILED. You need to give the agent concise, actionable feedback about what went wrong so it can try again.

Below is your internal knowledge about the task:

{context}

{file_diff_section}

RULES FOR FEEDBACK:
1. Be specific about what is wrong -- mention exact values, file names, or formats that are incorrect.
2. Do NOT give step-by-step instructions. Just point out the problem.
3. Do NOT reveal evaluator function names or code.
4. Keep it to 2-4 sentences.
5. If you can identify the specific mismatch between what the agent produced and what you expected, state it clearly.
\end{lstlisting}
\label{prompt:usersim:feedback}

\subsubsection{User Simulator: Context}
\begin{lstlisting}[style=prompt,caption={User Simulator Context},label={lst:usersim-context},numbers=none]
The {context} variable contains:

## Task Configuration
```json
{full task JSON config including instruction, evaluator spec, setup config}
```

## Evaluator Function
```python
{evaluator function source code found by AST-walking desktop_env/evaluators/}
```

## Input/Expected Files
{file contents: images as base64 image_url blocks, text files in full, spreadsheets as CSV, files >10MB as metadata only}

## Agent's trajectory
{step-by-step actions with screenshots from agent.executor.screenshot_inputs (last 40)}
\end{lstlisting}
\label{prompt:usersim:context}

\subsubsection{Binary Retry Signal}
\begin{lstlisting}[style=prompt,caption={Binary Retry Signal (injected after failed evaluation)},label={lst:binary-retry},numbers=none]
Your previous attempt did not succeed. The task is not complete. Please try again.
\end{lstlisting}
\label{prompt:binary:retry}

\subsubsection{Clarify Upon Retry Signal}
\begin{lstlisting}[style=prompt,caption={Clarify Upon Retry Signal (injected after failed evaluation)},label={lst:clarify-retry},numbers=none]
Your previous attempt did not succeed. Here is feedback from the user:
{feedback from user simulator}
\end{lstlisting}
\label{prompt:clarify:retry}

\subsection{Plan Extraction and Refinement}

\subsubsection{Plan Extraction and Refinement Base Prompt}
\label{app:pe_pr:base_prompt}
\begin{lstlisting}[style=prompt,caption={Plan Extraction and Refinement Prompt.},label={lst:sys-prompt},numbers=none]
You are analyzing agent trajectories to provide reflexion feedback for improving future performance.

You are given multiple rollouts from the SAME policy on the SAME task:
- Some SUCCESSFUL rollouts (score > 0)  
- Some FAILED rollouts (score = 0)

{PREVIOUS_FEEDBACK_SECTION}

For each rollout, you will receive:
1. **Visual Changes**: Fact captions describing what changed between screenshots at each step
2. **Agent Reasoning**: The agent's internal thoughts and decision-making process at each step

Your goal is to identify what worked in successful runs versus what went wrong in failed runs, then provide actionable feedback for future attempts.

ANALYSIS APPROACH:

1. **Compare Trajectories Step-by-Step**
   - Compare both visual changes and agent reasoning between successful/failed runs
   - Identify where successful and failed runs diverge in actions AND thinking
   - Note different approaches taken by successful vs failed runs
   - Look for consistent patterns across successes and failures

2. **Key Success Factors**
   - What specific actions or strategies led to success?
   - What reasoning patterns did successful runs exhibit?
   - Were there critical steps that successful runs handled correctly?
   - What environmental awareness did successful runs demonstrate?

3. **Failure Analysis**  
   - What mistakes did failed runs make in actions or reasoning?
   - Were there flawed reasoning patterns or incorrect decision making?
   - Did failed runs misunderstand the task or environment?
   - Were there missed opportunities or incorrect decisions?

4. **Actionable Insights**
   - What should the agent think about differently next time?
   - What should the agent do differently next time?
   - What should the agent avoid based on failed attempts?
   - Are there general principles or heuristics that emerge?

OUTPUT FORMAT:

<analysis>
[Detailed step-by-step comparison of successful vs failed trajectories, analyzing both visual changes and agent reasoning, identifying key divergence points and patterns]
</analysis>

<feedback>
DO:
1. [Specific actionable instruction based on successful patterns]
2. [Another positive action or strategy to follow]
3. [Key behavior or approach to adopt]
...
(up to a max of 10 items and can stop earlier if not applicable)

DON'T:
1. [Specific mistake or behavior to avoid from failed runs]
2. [Common error pattern to prevent]
3. [Action or thinking pattern that leads to failure]
...
(up to a max of 10 items and can stop earlier if not applicable)

PLAN:
[Step-by-step execution plan extracted from successful runs - concrete sequence of actions that leads to success]
</feedback>
\end{lstlisting}

\subsubsection{Plan Refinement and Addon Prompt}
\label{app:pe_pr:refinement_addon}
\begin{lstlisting}[style=prompt,caption={Plan Refinement Addon Prompt (PREVIOUS\_FEEDBACK\_SECTION)},label={lst:sys-prompt},numbers=none]
PREVIOUS FEEDBACK:
```
{previous_feedback}
```

ANALYSIS GOAL:
The previous feedback was applied, but the results show it may need refinement. You need to:
1. Identify which parts of the previous feedback were helpful vs harmful
2. Understand why the feedback led to the current outcomes  
3. Refine the feedback based on new evidence
\end{lstlisting}

\subsubsection{Plan Extraction with Historical Success}
\label{app:pe_pr:extract_historical_success}
\begin{lstlisting}[style=prompt,caption={Plan Extraction with Historical Success},label={lst:sys-prompt},numbers=none]
You are analyzing failed agent trajectories against a successful historical reference to generate corrective feedback.

SITUATION:
All current attempts failed, but we have a successful run from a previous trial to learn from.

HISTORICAL SUCCESSFUL RUN:
{historical_success_info}

CURRENT FAILED ATTEMPTS:
You will be shown multiple failed attempts from the current trial.

ANALYSIS GOAL:
Compare the failed attempts against the historical success to identify:
1. What the successful run did right that current attempts are missing
2. What systematic errors current attempts are making
3. How to guide future attempts toward the successful pattern

For each run, you will see:
1. **Visual Changes**: Fact captions describing what changed between screenshots at each step  
2. **Agent Reasoning**: The agent's internal thoughts and decision-making process at each step

ANALYSIS APPROACH:

1. **Success Pattern Analysis**
   - What made the historical run successful?
   - What key decisions, actions, or reasoning patterns led to success?
   - What environmental awareness did it demonstrate?

2. **Current Failure Analysis**
   - How do current failures differ from the successful pattern?
   - What systematic mistakes are being repeated?
   - Where do current attempts go wrong compared to the success?

3. **Corrective Strategy**
   - How can future attempts align with the successful pattern?
   - What specific changes in reasoning or actions are needed?
   - What should be avoided based on current failures?

OUTPUT FORMAT:

<analysis>
[Detailed comparison of historical success vs current failures, identifying key differences and systematic error patterns]
</analysis>

<feedback>
DO:
1. [Key strategy from the successful run to adopt]
2. [Specific action or approach that led to success]
3. [Reasoning pattern that worked in the successful case]
...
(up to a max of 10 items and can stop earlier if not applicable)

DON'T:
1. [Specific mistake in current attempts to avoid]
2. [Error pattern that differs from successful approach]
3. [Action or thinking that leads to failure vs success]
...
(up to a max of 10 items and can stop earlier if not applicable)

PLAN:
[Step-by-step execution plan extracted from the historical successful run - concrete sequence of actions to replicate]
</feedback>
\end{lstlisting}

\subsubsection{Plan Extraction from Failures}
\label{app:pe_pr:extract_all_failures}
\begin{lstlisting}[style=prompt,caption={Plan Extraction from Failures},label={lst:sys-prompt},numbers=none]
You are analyzing failed agent trajectories to generate diagnostic feedback when no successful examples exist.

SITUATION:
All attempts have failed, and there are no successful runs from previous trials to reference.

CURRENT FAILED ATTEMPTS:
You will be shown multiple failed attempts. Your goal is to identify the most promising approach and provide guidance to help future attempts succeed.

ANALYSIS GOAL:
Since all attempts failed, focus on:
1. Which failure showed the most promise/progress
2. What systematic errors are preventing success
3. What fundamental strategies might lead to success

For each failed run, you will see:
1. **Visual Changes**: Fact captions describing what changed between screenshots at each step
2. **Agent Reasoning**: The agent's internal thoughts and decision-making process at each step

ANALYSIS APPROACH:

1. **Comparative Failure Analysis**
   - Which attempt made the most progress before failing?
   - What different approaches were tried?
   - What common error patterns appear across attempts?

2. **Root Cause Identification**
   - What fundamental issues are causing failures?
   - Are failures due to misunderstanding, poor execution, or environment issues?
   - What capabilities seem to be missing?

3. **Diagnostic Strategy**
   - What should future attempts focus on first?
   - What basic principles or approaches might help?
   - What specific errors should be avoided?

OUTPUT FORMAT:

<analysis>
[Detailed analysis of failure patterns, identifying the most promising approach and systematic issues preventing success]
</analysis>

<feedback>
DO:
1. [Basic strategy that might lead to success]
2. [Fundamental approach to try based on most promising failure]
3. [Key capability or awareness to develop]
...
(up to a max of 10 items and can stop earlier if not applicable)

DON'T:
1. [Common error pattern to avoid across all attempts]
2. [Systematic mistake that prevents success]
3. [Flawed approach or reasoning to prevent]
...
(up to a max of 10 items and can stop earlier if not applicable)
</feedback>
\end{lstlisting}

\subsubsection{Planning Feedback Addon}
\label{app:pe_pr:plan_feedback_addon}
\begin{lstlisting}[style=prompt,caption={Planning Feedback Addon},label={lst:sys-prompt},numbers=none] [REFLECTION FEEDBACK] LEARNING FROM PREVIOUS ATTEMPTS:

Task: `{instruction}`

Based on analysis of previous attempts at this task, here's what you should know:

{feedback_text}

IMPORTANT: Use this feedback to guide your approach. Pay attention to what has worked and what has failed in similar situations. Adapt your strategy based on these insights while staying focused on completing the task successfully.
\end{lstlisting}

\clearpage
\section{Qualitative Analysis}

\subsection{Clarified Instruction Examples}
\label{app:clarification:qualitative}

We present qualitative examples illustrating the effects of instruction clarification. Each example shows the original instruction and task context, the LLM-generated clarified instruction, and (when applicable) a human-corrected version. We highlight both successful clarifications that improve alignment with the evaluator, as well as failure cases where clarification introduces unintended artifacts such as impossible constraints or trivial solutions.
\vspace{-2em}
\begin{center}
\rule{\linewidth}{0.4pt}
\end{center}
\vspace{-2em}

\subsection*{Example 1: Improved evaluator alignment}

\noindent\textbf{Original Instruction:} \textit{Append one entry of AI researcher Yann LeCun from Google Scholar into an existing table researchers.xlsx.}

\noindent\textbf{Clarification Context}
\begin{itemize}
    \item Original \texttt{researchers.xlsx} file with formatting
    \item Evaluation JSON with arguments to evaluator function \texttt{literal\_match}
    \item Implementation of \texttt{literal\_match} function
    \item Initial screenshot (google.com)
\end{itemize}

\noindent\textbf{Clarified Instruction (LLM)}
\vspace{-2em}
\begin{tcolorbox}
[colback=blue!5!white,colframe=blue!50!black]
\small
Open \texttt{/home/user/Desktop/researchers.xlsx} and append a single new row at the end for AI researcher Yann LeCun, using his Google Scholar profile at \texttt{https://scholar.google.com/citations?user=WLN3QrAAAAAJ\&hl=en}. Fill the same six columns in this order—Name, All citations, h-index, i10-index, Top Cited Work, PDF Link of Top Cited Work—using the “All” values for the three metrics; for Top Cited Work, use the title of the first listed publication on his profile, and for the PDF link, open that publication and copy the external PDF/full-text link shown on its details page. Save the file.
\end{tcolorbox}

\vspace{0.5em}

\noindent\textbf{Analysis}
\begin{tcolorbox}[colback=gray!5!white,colframe=gray!50!black]
\small
The clarification directly aligned the agent with the evaluator’s exact-matching requirements:

\begin{itemize}
    \item \textbf{File specification:} Provides the exact file path and explicitly instructs appending a single new row, ensuring correct list length and ordering.
    \item \textbf{Source grounding:} Specifies the exact Google Scholar profile URL, avoiding ambiguity and failure modes such as search errors, CAPTCHA, or login interruptions.
    \item \textbf{Schema alignment:} Defines the six required columns and their order, and specifies use of the “All” metrics, preventing mismatches in format or subset selection.
    \item \textbf{Deterministic extraction:} Instructs selecting the first listed publication and extracting the external PDF/full-text link from the article page, leading to consistent outputs.
    \item \textbf{Completion requirement:} Explicitly instructs saving the file, ensuring the final state matches evaluator expectations.
\end{itemize}

In the original instruction, agents frequently failed to append a row, produced incomplete or incorrect fields (e.g., wrong paper or link), or became stuck during web navigation—all of which result in failure under \texttt{literal\_match}. After clarification, agent behavior becomes consistent with the evaluator’s requirements, improving success from 0\% to 100\%.
\end{tcolorbox}


\clearpage

\subsection*{Example 2: Instruction - evaluator mismatch}

\noindent\textbf{Original Instruction:} \textit{I am a Chinese citizen and I want to go to Macau to watch a concert recently, but I have not yet applied for a visa for Macau. I live in Futian District, Shenzhen City. I heard that Shenzhen currently has 24-hour self-service check-in machines. Please help me find the addresses of 5 24-hour self-service check-in machines in Futian District and save them in Chinese in this open word document.}

\noindent\textbf{Clarification Context}
\begin{itemize}
    \item Original \texttt{AllLocations.docx} file
    \item Evaluation JSON containing potential match arguments to evaluator function \texttt{fuzzy\_place\_math}
    \item Implementation of \texttt{fuzzy\_place\_math} function (incorrectly implemented, shown below)

\begin{tcolorbox}[colback=black!2!white,colframe=black!50!black]
\tiny
\begin{verbatim}
def fuzzy_place_math(result_file_path, rules) -> float:
    if result_file_path is None:
        return 0.
    expect = rules["expected"]  # a list of possible answers

    doc = Document(result_file_path)
    words_list = []
    for para in doc.paragraphs:
        words_list.extend(para.text.split())

    fuzzy_score_list = []
    for word in words_list:
        max_score = 0
        for ans in expect:
            score = fuzz.ratio(word, ans) / 100
            max_score = max(max_score, score)
        fuzzy_score_list.append(max_score)
    ##################################################################
    # BUG: requires exactly 3 addresses, otherwise automatic failure #
    ##################################################################
    if len(fuzzy_score_list) != 3:
        return 0.

    return sum(fuzzy_score_list) / 3
\end{verbatim}
\end{tcolorbox}
    \item Initial screenshot (empty \texttt{AllLocations.docx} file)
\end{itemize}

\noindent\textbf{Clarified Instruction (LLM)}
\begin{tcolorbox}[colback=blue!5!white,colframe=blue!50!black]
\small
I’m a Chinese citizen planning to go to Macau for a concert and haven’t applied for the visa yet. I live in Futian District, Shenzhen. I heard there are 24-hour self-service check-in machines in Shenzhen. Please help me find three (3) addresses of 24-hour self-service check-in machines located in Futian District and put them into the currently open document named \texttt{AllLocations.docx} on my Desktop.  Please write the addresses in Simplified Chinese, with each address alone on its own line, and don’t add any other text, numbering, bullets, or extra spaces before/after. The document should contain only these three lines. Save the changes to the same file.
\end{tcolorbox}

\noindent\textbf{Analysis}
\begin{tcolorbox}[colback=gray!5!white,colframe=gray!50!black]
\small
The clarification changed the required output from “5 addresses” to exactly three addresses and added strict formatting constraints (only three lines, no extra text/numbering/bullets, no extraneous spaces, save to the same file). The evaluator demands exactly 3 whitespace-separated tokens in \texttt{AllLocations.docx}; any additional tokens cause an automatic fail. Under the original instruction, agents included a title and five items (often with spaces), producing more than three tokens and yielding 0/3 successes. After clarification, agents produced exactly three standalone address lines and removed extras, aligning with the evaluator’s len==3 check and enabling high fuzzy matches to the expected answers, resulting in 3/3 successes. This is a clear, instruction-driven alignment improvement rather than random variance.
\end{tcolorbox}


\clearpage

\subsection*{Example 3: Human-corrected impossible task after clarification}

\noindent\textbf{Original Instruction:} \textit{I've noticed that the image on the second slide is too dim. Can you please enhance its brightness for me? Save the adjusted image on the Desktop and name it "background.png". Thank you!}

\vspace{0.3em}
\noindent\textbf{Clarification Context}
\begin{itemize}
    \item Original \texttt{PPT-Template\_widescreen.pptx} file with formatting
    \item Evaluation JSON with arguments to evaluator function \\ \texttt{check\_brightness\_decrease\_and\_structure\_sim}
    \item Implementation of \texttt{check\_brightness\_decrease\_and\_structure\_sim} function
    \item Initial screenshot (opened \texttt{PPT-Template\_widescreen.pptx} file)
\end{itemize}

\vspace{0.5em}
\noindent\textbf{Clarified Instruction (LLM)}
\begin{tcolorbox}[colback=red!5!white,colframe=red!50!black]
\small
On the second slide, the image looks a bit too bright—could you slightly reduce its brightness so it’s darker, but keep everything else the same (no cropping, resizing, or other edits)? When you’re done, please export just that image as a PNG and save it on the Desktop as \texttt{/home/user/Desktop/background.png}. Thanks!
\end{tcolorbox}

\noindent\textbf{Issue:}
The clarified instruction inverts the original task due to the confusing evaluation function name; the evaluation JSON checks that the original photo is dimmer than the new photo rather than checking that the new photo is brighter than the original.

\vspace{0.3em}
\noindent\textbf{Human-Corrected Instruction:}
\begin{tcolorbox}[colback=green!5!white,colframe=green!50!black]
\small
On the second slide, the image looks a bit too dim—could you slightly enhance its brightness so it’s brighter, but keep everything else the same (no cropping, resizing, or other edits)? When you’re done, please export just that image as a PNG and save it on the Desktop as \texttt{/home/user/Desktop/background.png}. Thanks!
\end{tcolorbox}


\clearpage

\subsection*{Example 4: Human-corrected trivial task after clarification}

\noindent\textbf{Original Instruction:} \textit{Find discussions of community and open one with most replies.}

\vspace{0.3em}
\noindent\textbf{Clarification Context}
\begin{itemize}
    \item Evaluation JSON with arguments to evaluator function \texttt{is\_expected\_active\_tab}
    \item Implementation of \texttt{is\_expected\_active\_tab} function
    \item Initial screenshot (google.com)
\end{itemize}

\vspace{0.5em}
\noindent\textbf{Clarified Instruction (LLM)}
\begin{tcolorbox}[colback=red!5!white,colframe=red!50!black]
\small
Go to FlightAware’s Community/Discussions forum and open the topic with the most replies—The Banter Thread—so that the active tab is https://discussions.flightaware.com/t/the-banter-thread/4412.
\end{tcolorbox}

\noindent\textbf{Issue:}
The clarified instruction includes the ground-truth URL that is checked which trivializes the search process in the task.

\vspace{0.3em}
\noindent\textbf{Human-Corrected Instruction:}
\begin{tcolorbox}[colback=green!5!white,colframe=green!50!black]
\small
Go to FlightAware’s Community/Discussions forum and open the topic with the most replies on the entire website.
\end{tcolorbox}

\clearpage
\subsection{Retry (Clarify) with User Simulator}
\label{app:user_simulator:qualitative}

We present three qualitative examples (one per model) where the user simulator's feedback resolved a genuine ambiguity in the task instruction, enabling the agent to recover from failure. In each case, the agent's first interpretation was reasonable but did not match the evaluator's expectation; the user simulator clarified the intended meaning without providing step-by-step instructions.

\begin{center}
\rule{\linewidth}{0.4pt}
\end{center}

\subsection*{Example 1: Ambiguous scope:document-level vs.\ global default (GPT-5)}

\noindent\textbf{Instruction:} \textit{``Make Times New Roman the default Font.''}\\
\noindent\textbf{Outcome:} Pass after 1 retry (32 total steps).

\vspace{0.5em}
\noindent\textbf{First attempt (failed):}
The instruction does not specify whether ``default'' means the default for the current document or the global LibreOffice Writer default for all future documents. The agent reasonably changed the Default Paragraph Style in the open document:a valid interpretation that only affects that file.

\vspace{0.3em}
\noindent\begin{tabular}{@{}ccc@{}}
\includegraphics[width=0.30\linewidth]{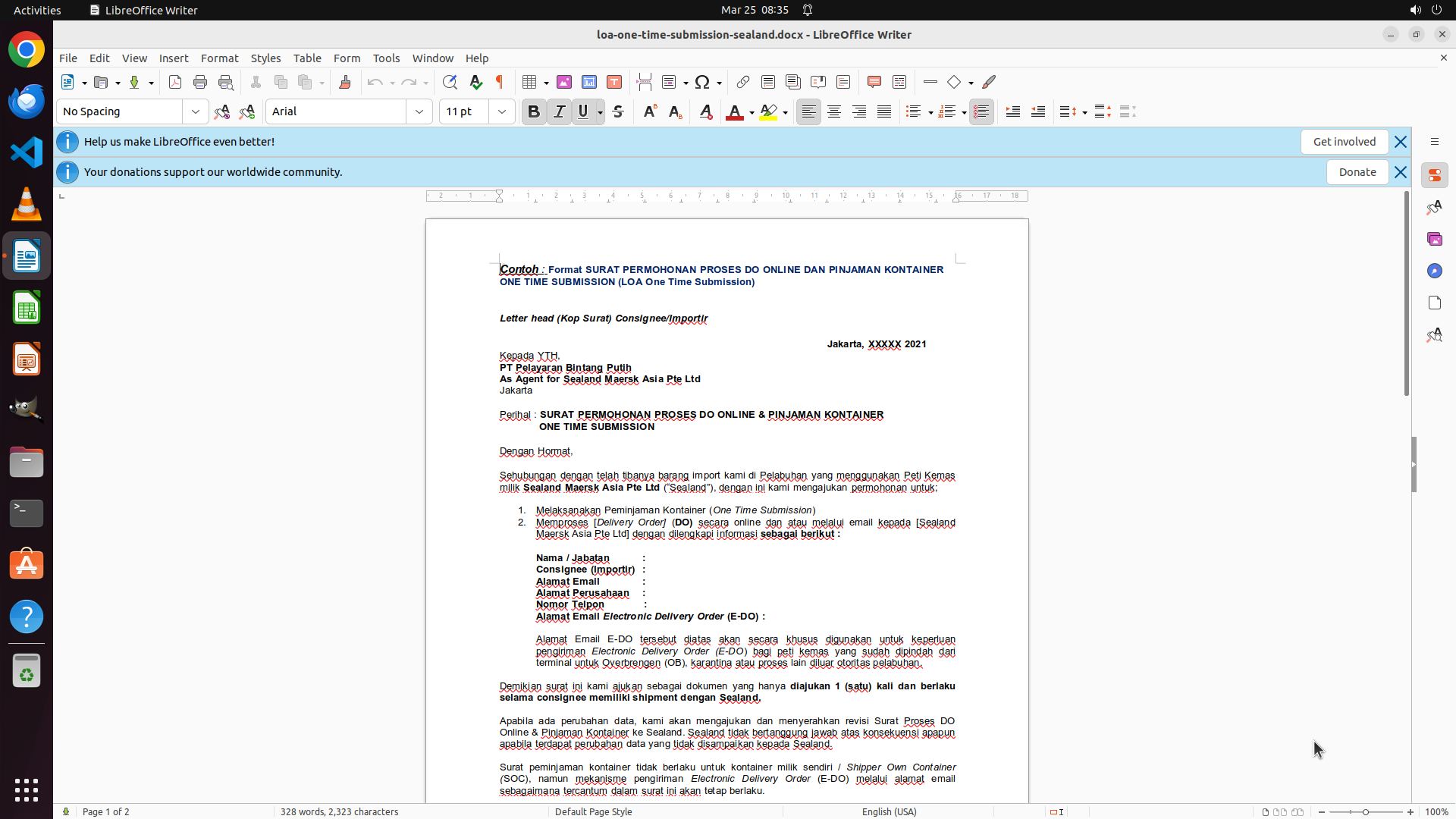} &
\includegraphics[width=0.30\linewidth]{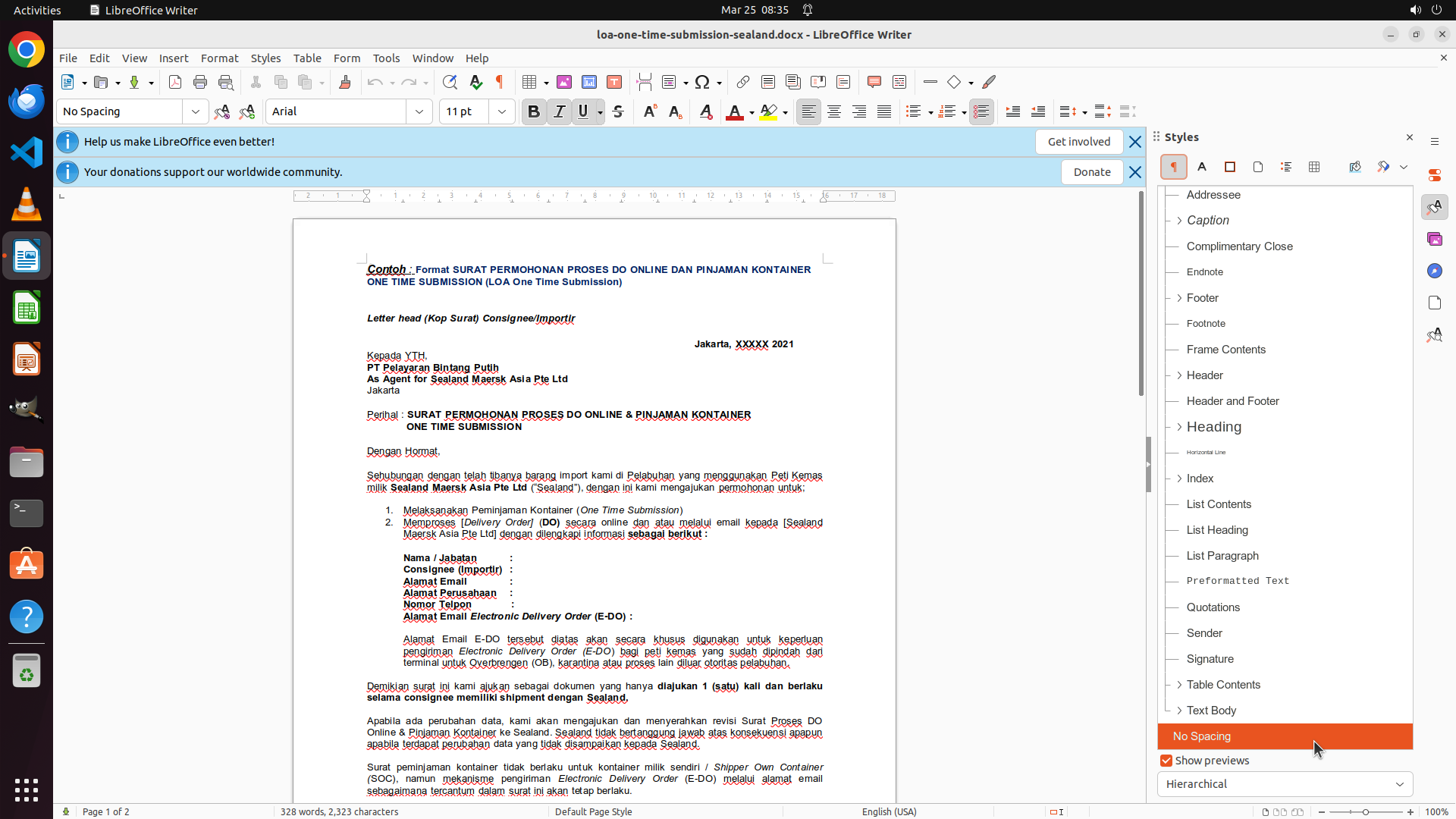} &
\includegraphics[width=0.30\linewidth]{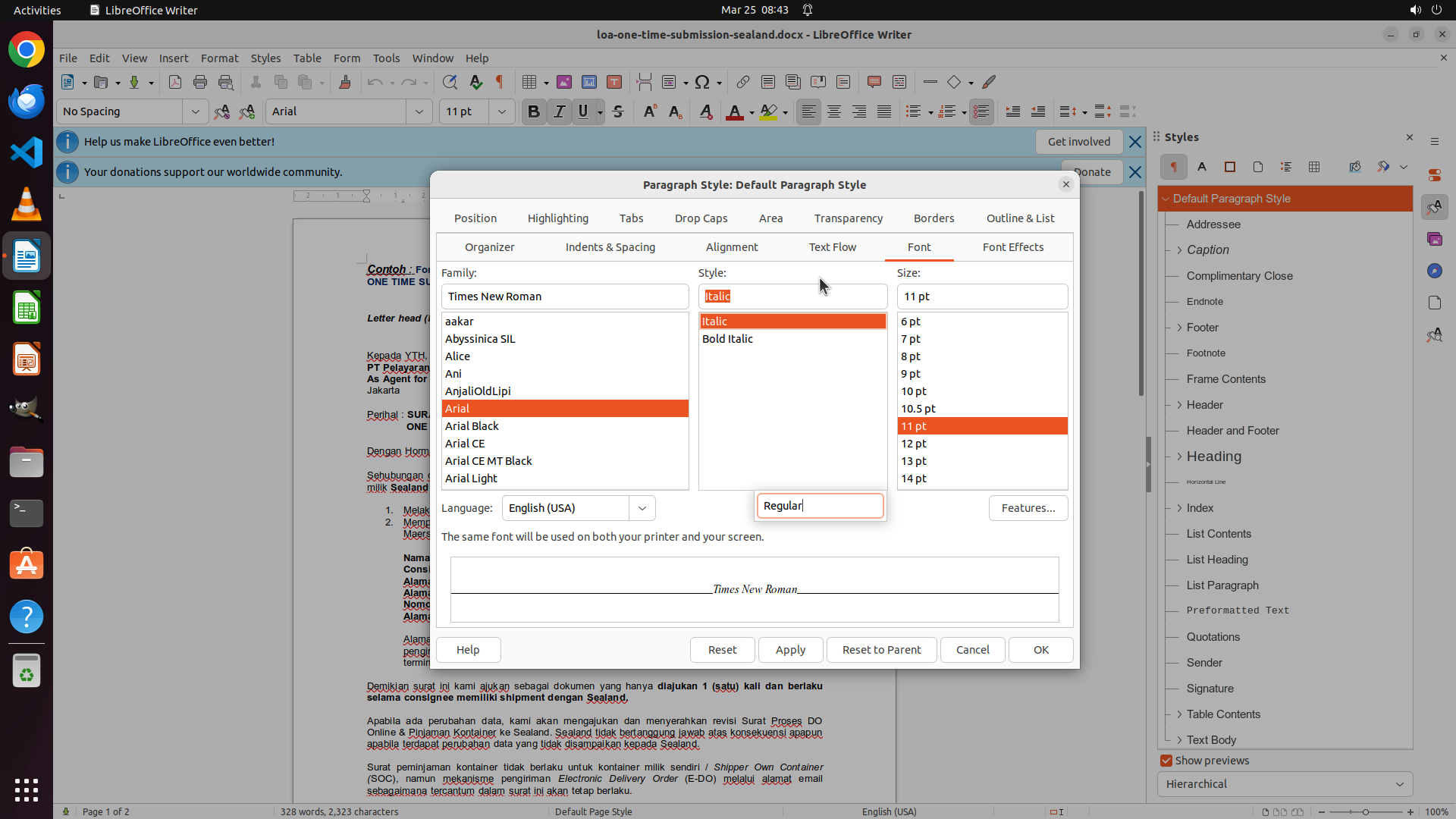} \\[-2pt]
{\scriptsize (a) Initial state} &
{\scriptsize (b) Open Styles sidebar (F11)} &
{\scriptsize (c) Set font in Paragraph Style dialog} \\
\end{tabular}

\begin{tcolorbox}[colback=red!5!white,colframe=red!50!black,title={\small User Simulator Feedback}]
\small You changed the ``Default Paragraph Style'' in the open document, which only affects that file. The global Writer default wasn't updated: in \texttt{\~{}/.config/libreoffice/.../registrymodifications.xcu}, the default font is not set to ``Times New Roman''. Update LibreOffice Writer's global ``Default'' font (Basic Fonts/Western) so that this key's value becomes ``Times New Roman''.
\end{tcolorbox}

\noindent\textbf{After feedback (succeeded):}
The user simulator clarified that ``default font'' means the \emph{global} setting, not the document-level style. The agent navigated to Tools~$\to$~Options~$\to$~Basic Fonts (Western) and updated the system-wide default.

\vspace{0.3em}
\noindent\begin{tabular}{@{}ccc@{}}
\includegraphics[width=0.30\linewidth]{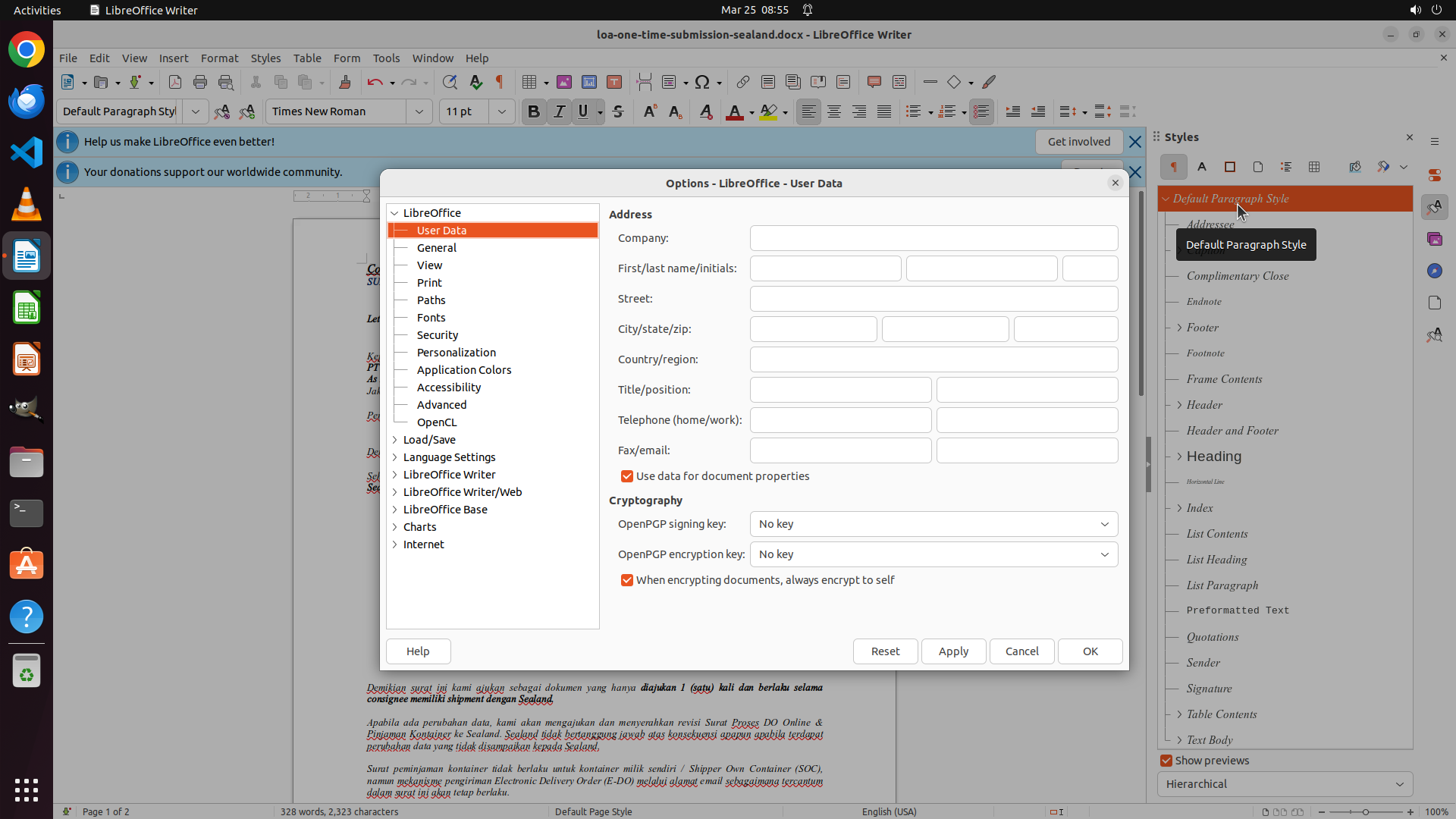} &
\includegraphics[width=0.30\linewidth]{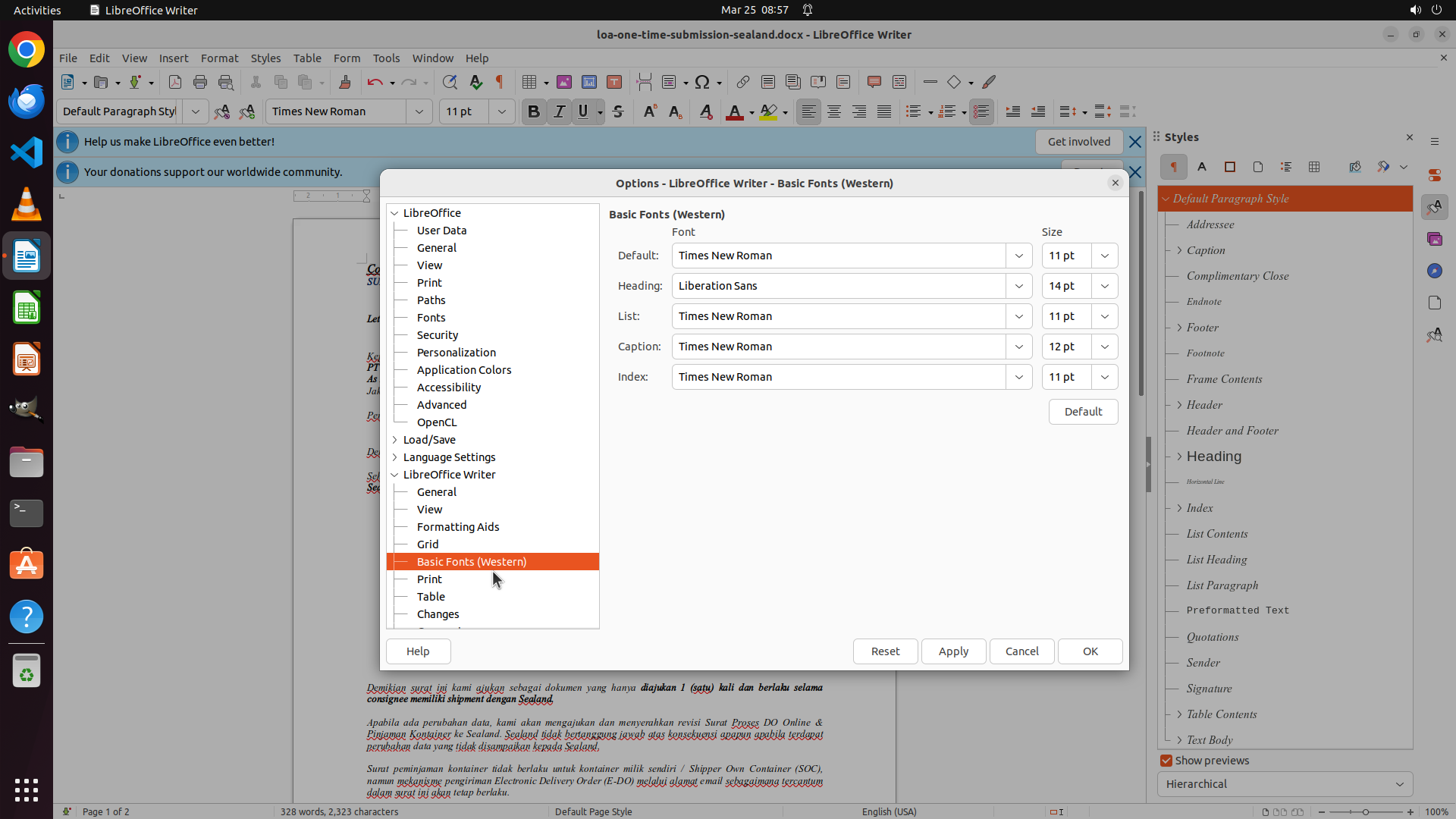} &
\includegraphics[width=0.30\linewidth]{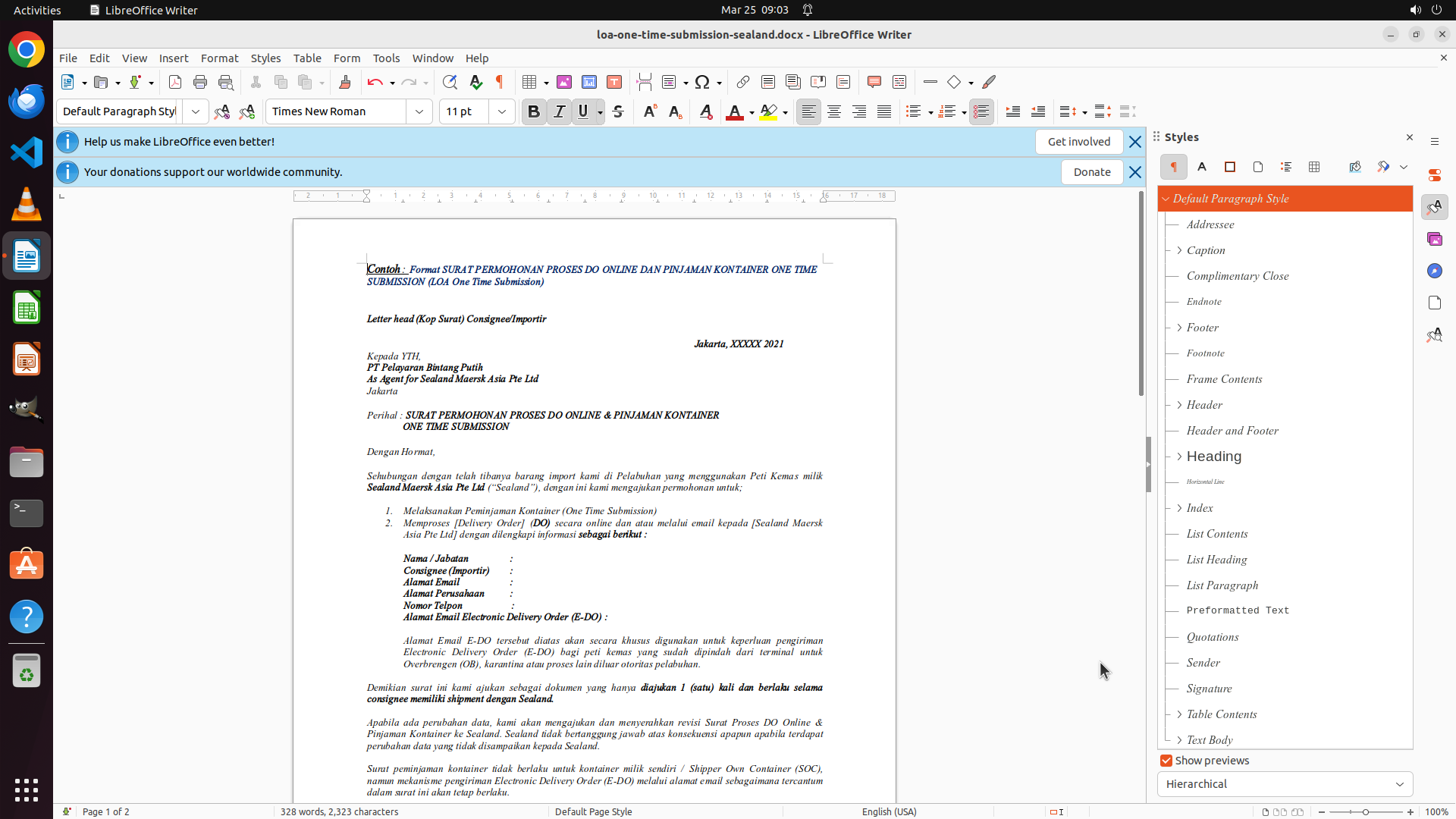} \\[-2pt]
{\scriptsize (d) Open Tools $\to$ Options} &
{\scriptsize (e) Set Basic Fonts (Western)} &
{\scriptsize (f) Confirm and call \texttt{done()}} \\
\end{tabular}


\clearpage
\subsection*{Example 2: Ambiguous setting name:\texttt{idle-dim} vs.\ \texttt{idle-delay} (Claude Sonnet 4.6)}

\noindent\textbf{Instruction:} \textit{``Could you set the `Dim screen when inactive' to off in setting?''}\\
\noindent\textbf{Outcome:} Pass after 1 retry (16 total steps).

\vspace{0.5em}
\noindent\textbf{First attempt (failed):}
The instruction references ``Dim screen when inactive,'' which maps naturally to the GNOME setting \texttt{idle-dim}. The agent opened Settings, then used the terminal to run \texttt{gsettings set org.gnome.settings-daemon.plugins.power idle-dim false}:a reasonable interpretation. However, the evaluator checks \texttt{org.gnome.desktop.session idle-delay}, which controls the inactivity timeout rather than the dimming behavior.

\vspace{0.3em}
\noindent\begin{tabular}{@{}ccc@{}}
\includegraphics[width=0.30\linewidth]{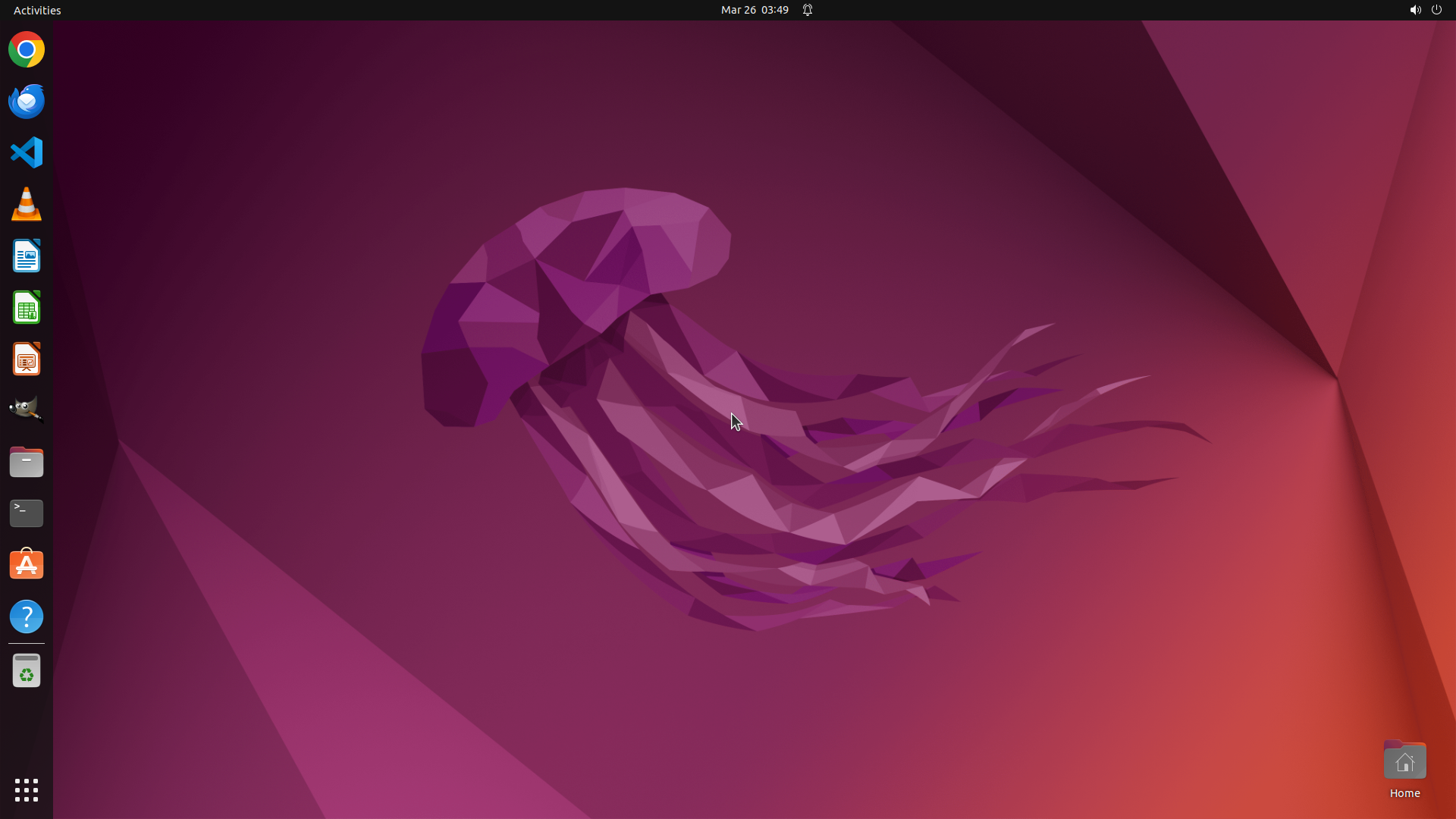} &
\includegraphics[width=0.30\linewidth]{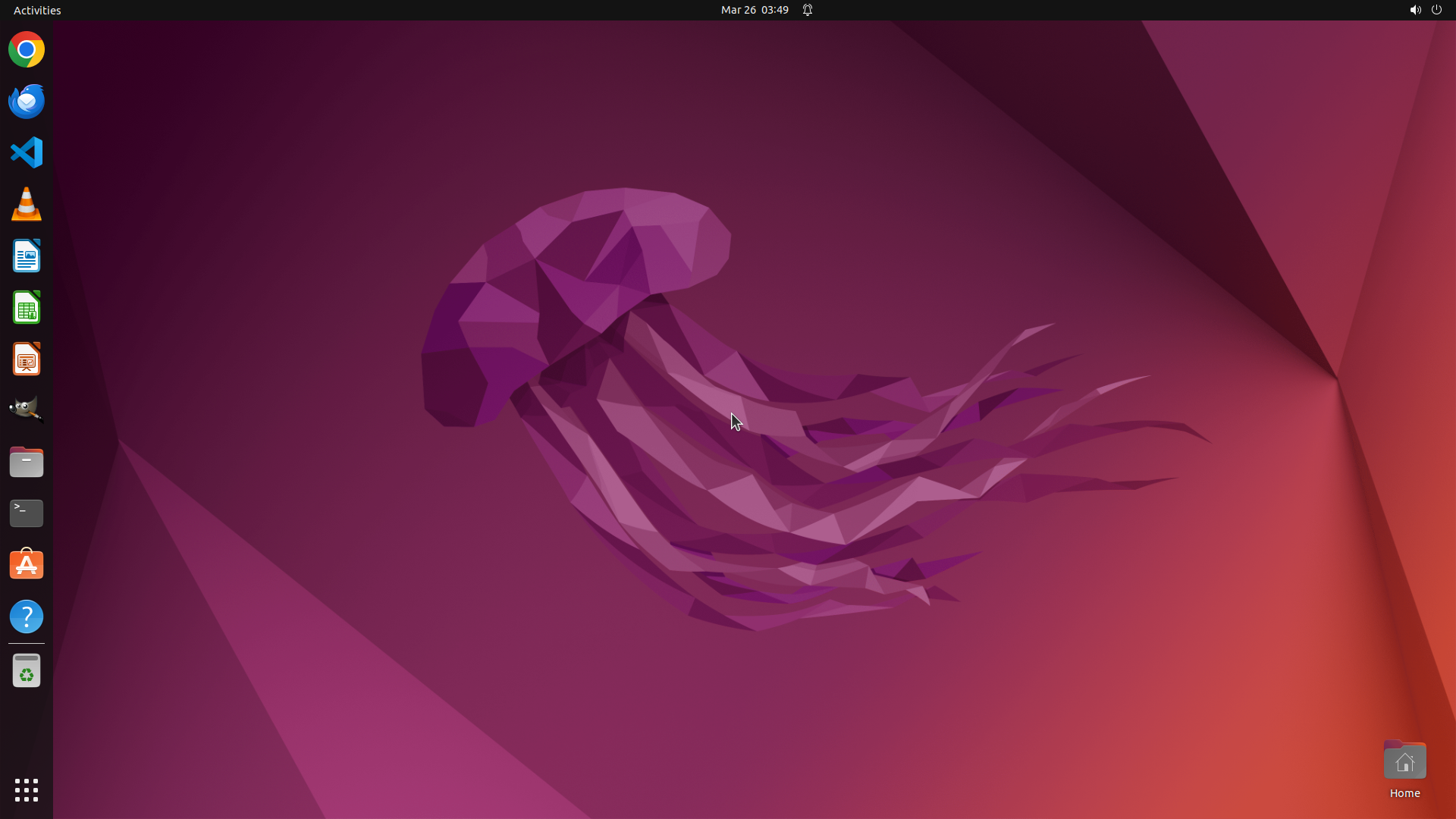} &
\includegraphics[width=0.30\linewidth]{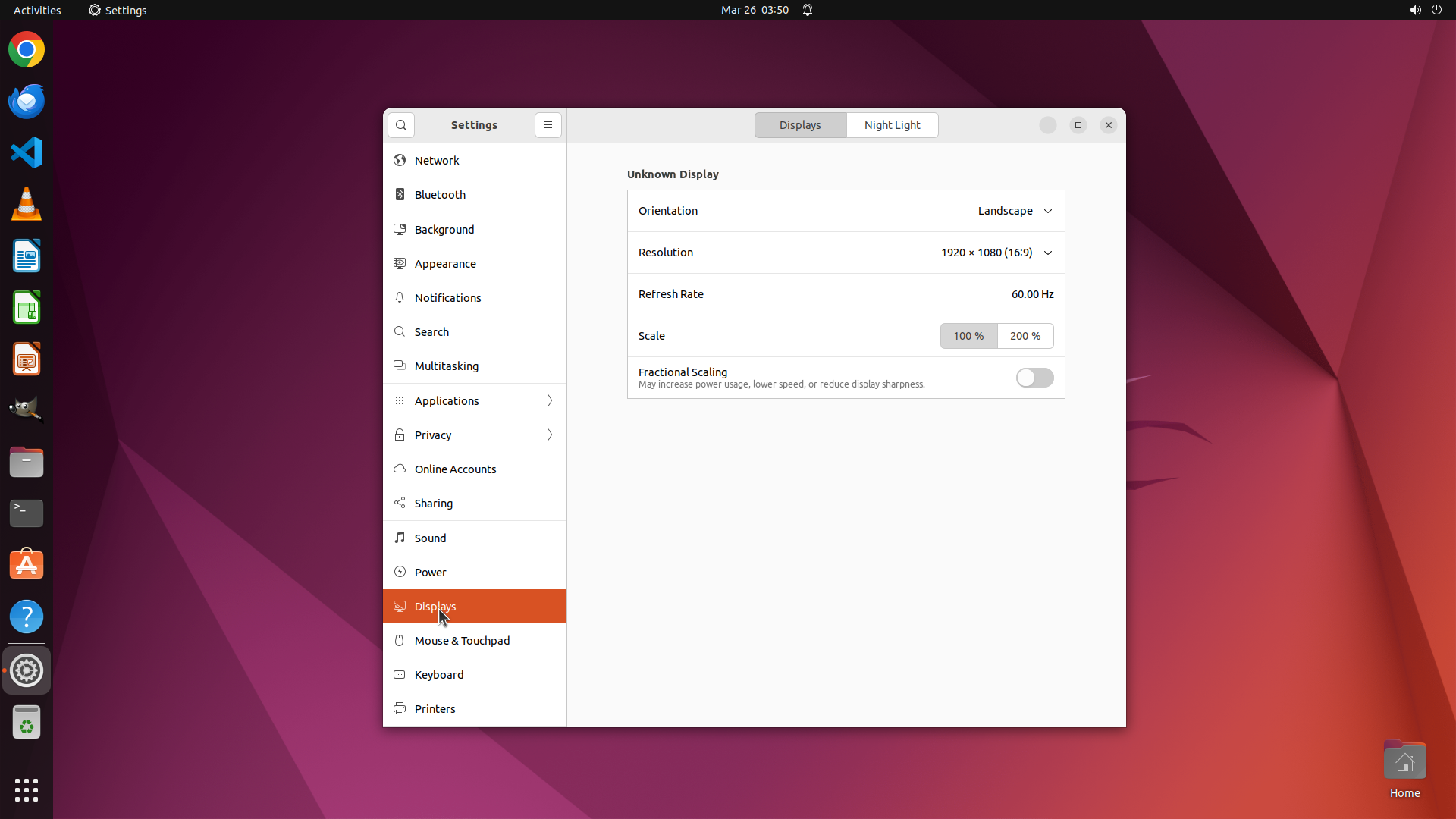} \\[-2pt]
{\scriptsize (a) Initial: Ubuntu desktop} &
{\scriptsize (b) Screenshot (no visible change)} &
{\scriptsize (c) Settings open on Display page} \\
\end{tabular}

\begin{tcolorbox}[colback=red!5!white,colframe=red!50!black,title={\small User Simulator Feedback}]
\small You changed the wrong setting. You set \texttt{org.gnome.settings-daemon.plugins.power idle-dim} to \texttt{false}, but the required key is \texttt{org.gnome.desktop.session idle-delay}, which must be \texttt{0} (expected output: ``uint32 0''). Because \texttt{idle-delay} wasn't set to 0, the check fails. Update \texttt{idle-delay} to 0.
\end{tcolorbox}

\noindent\textbf{After feedback (succeeded):}
The user simulator specified the exact gsettings key that needs to be changed. The agent ran \texttt{gsettings set org.gnome.desktop.session idle-delay 0} in the terminal.

\vspace{0.3em}
\noindent\begin{tabular}{@{}ccc@{}}
\includegraphics[width=0.30\linewidth]{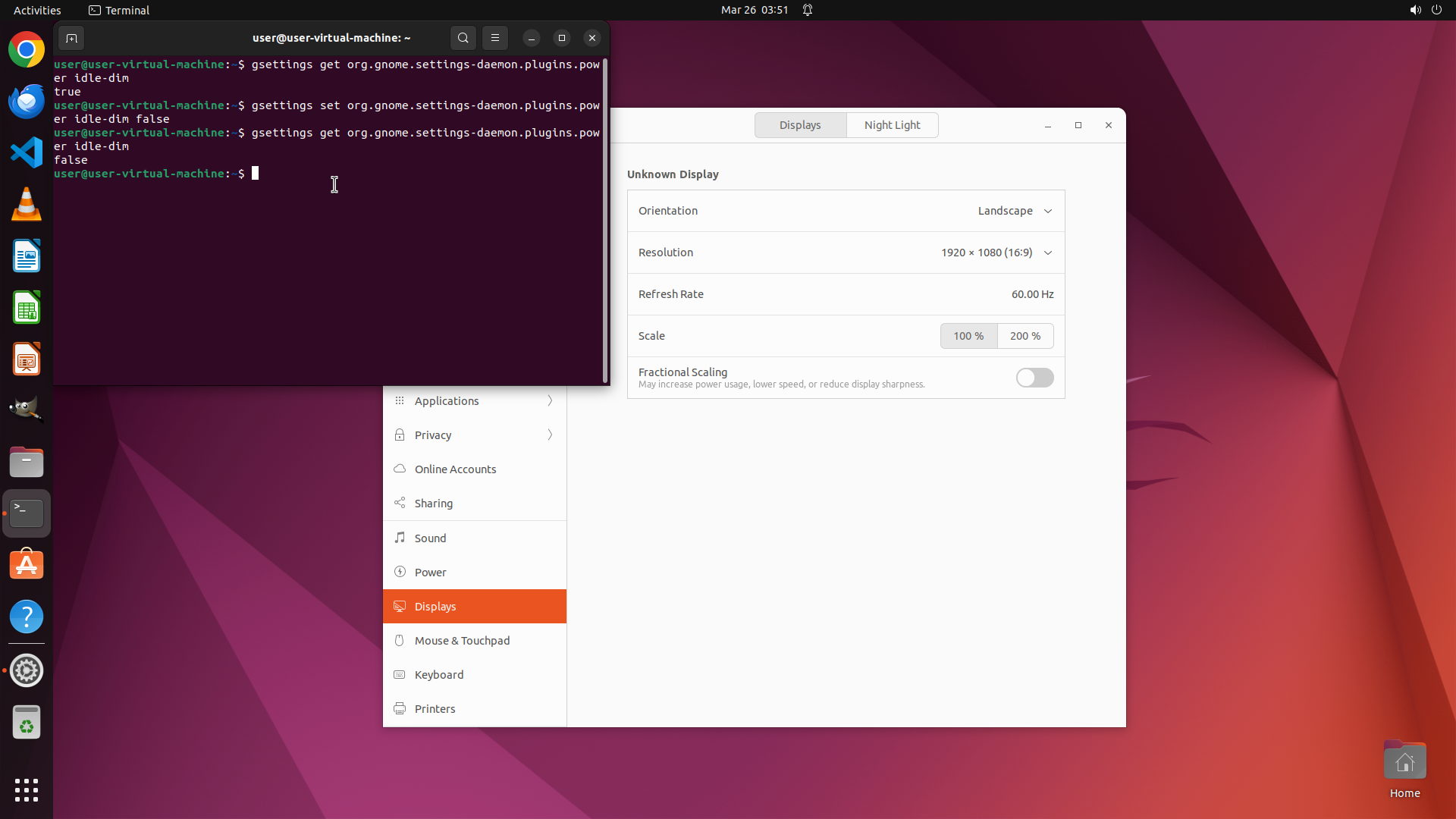} &
\includegraphics[width=0.30\linewidth]{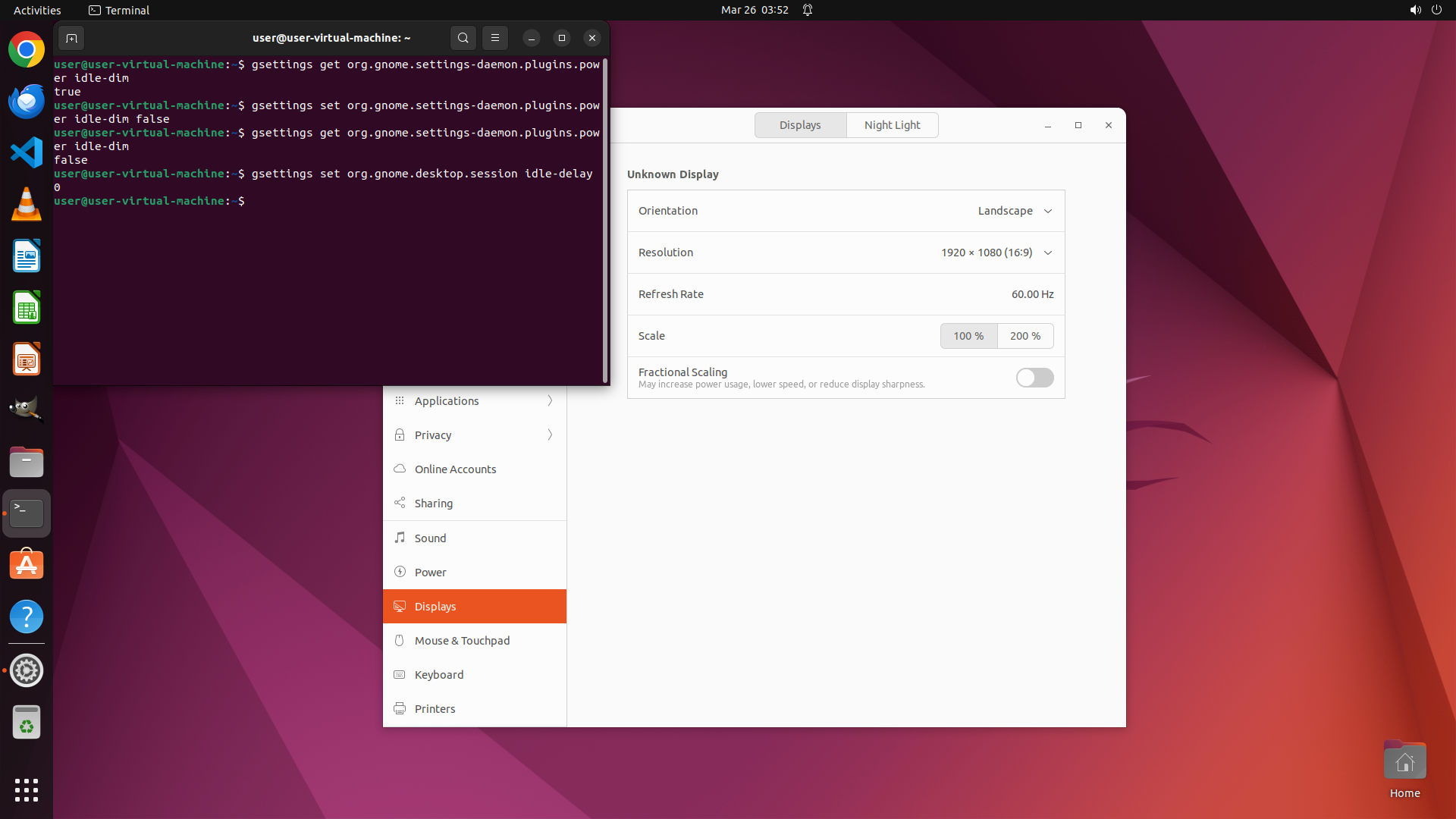} &
\includegraphics[width=0.30\linewidth]{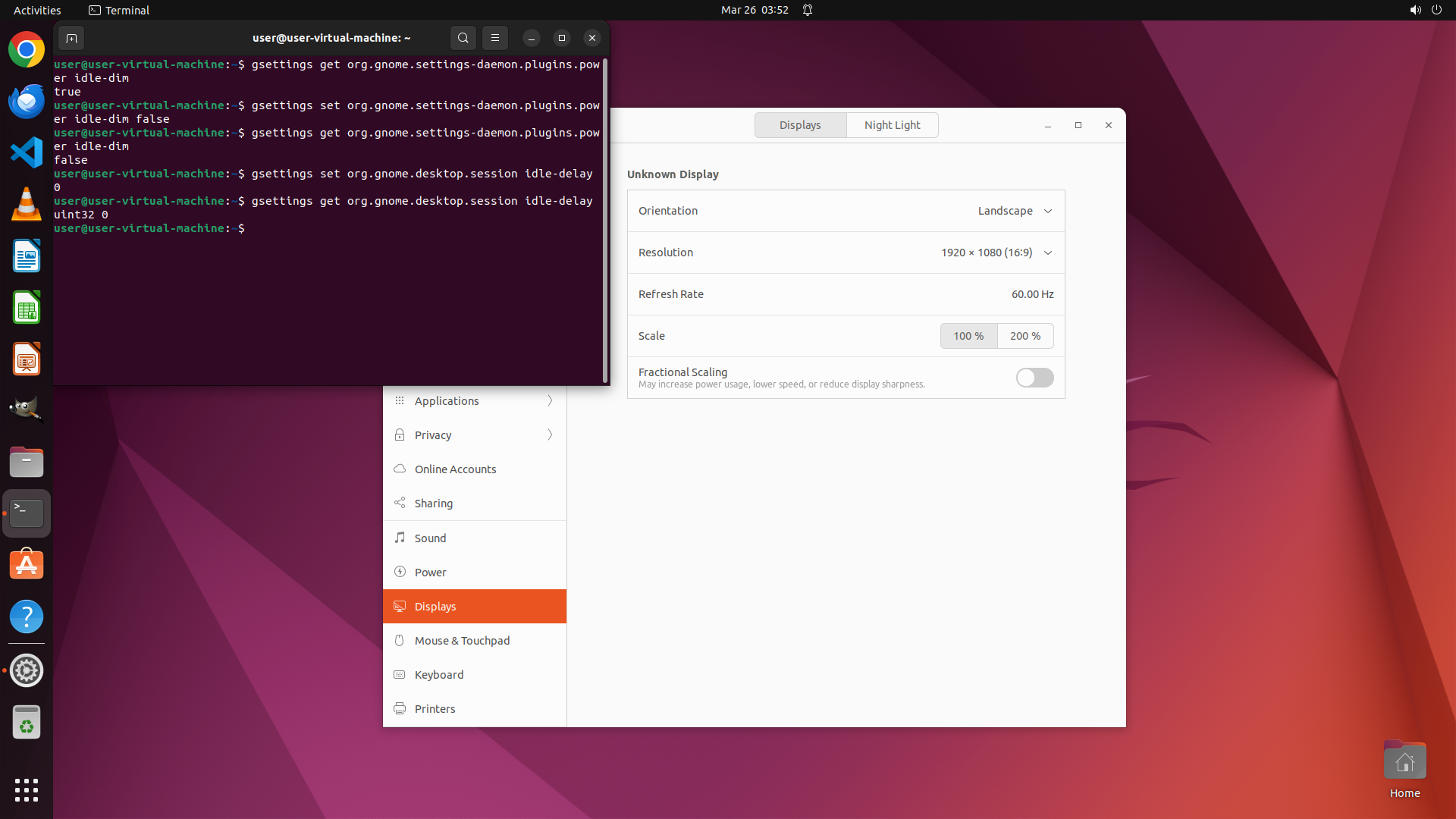} \\[-2pt]
{\scriptsize (d) Terminal: ran \texttt{idle-dim false}} &
{\scriptsize (e) Runs \texttt{gsettings set idle-delay 0}} &
{\scriptsize (f) Verifies \texttt{uint32 0}, calls \texttt{done()}} \\
\end{tabular}


\clearpage
\subsection*{Example 3: Ambiguous term:``note'' means speaker notes, not comment (Kimi K2.5)}

\noindent\textbf{Instruction:} \textit{``Add a note `APP' into the slide and give the slide a purple background color.''}\\
\noindent\textbf{Outcome:} Pass after 1 retry (16 total steps).

\vspace{0.5em}
\noindent\textbf{First attempt (failed):}
The word ``note'' in LibreOffice Impress is ambiguous: it could refer to a \emph{comment} (an annotation visible on the slide), a \emph{text box}, or \emph{speaker notes} (the notes pane below the slide used during presentations). The agent added a comment annotation:a reasonable interpretation. The evaluator expected speaker notes.

\vspace{0.3em}
\noindent\begin{tabular}{@{}ccc@{}}
\includegraphics[width=0.30\linewidth]{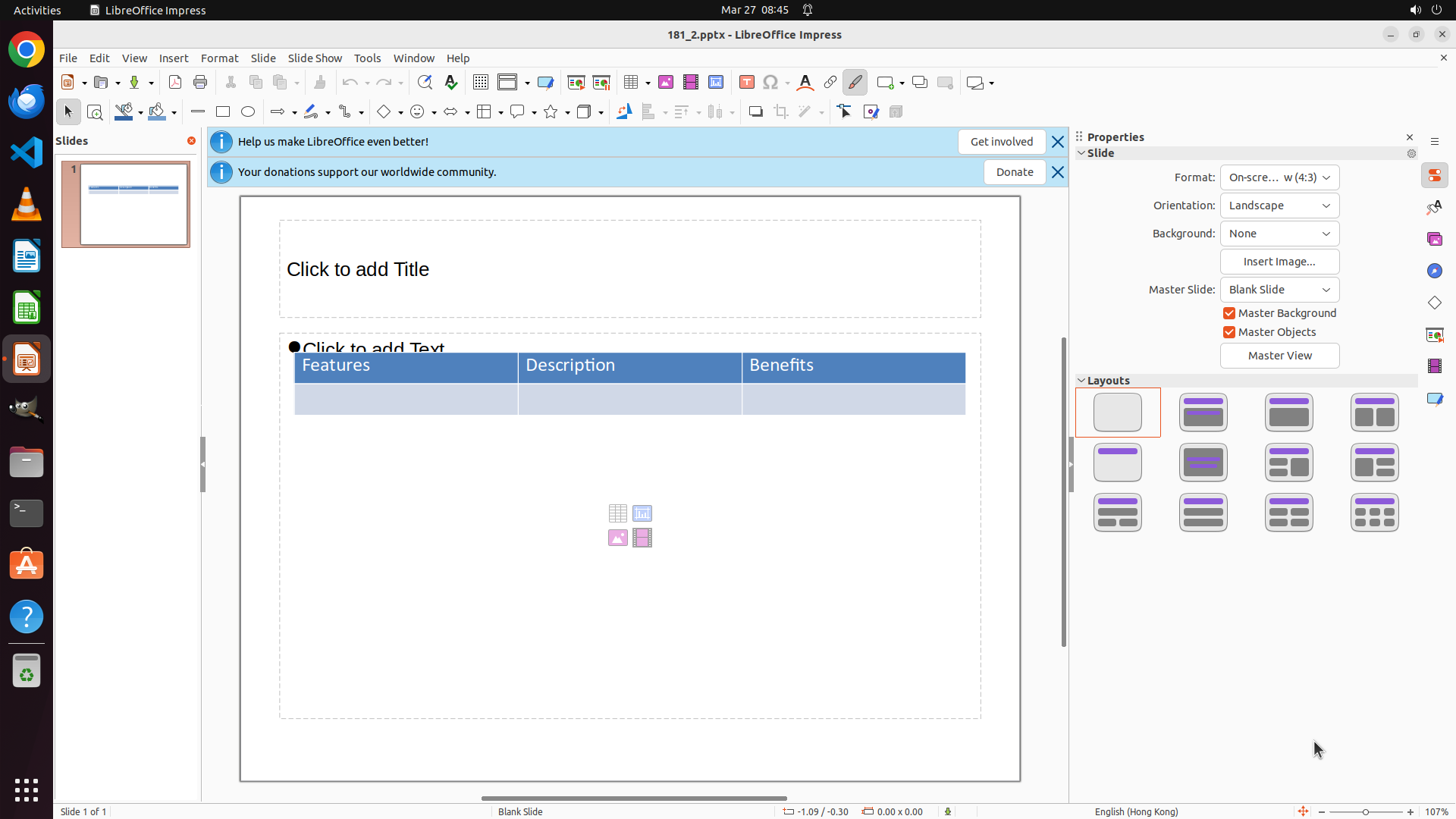} &
\includegraphics[width=0.30\linewidth]{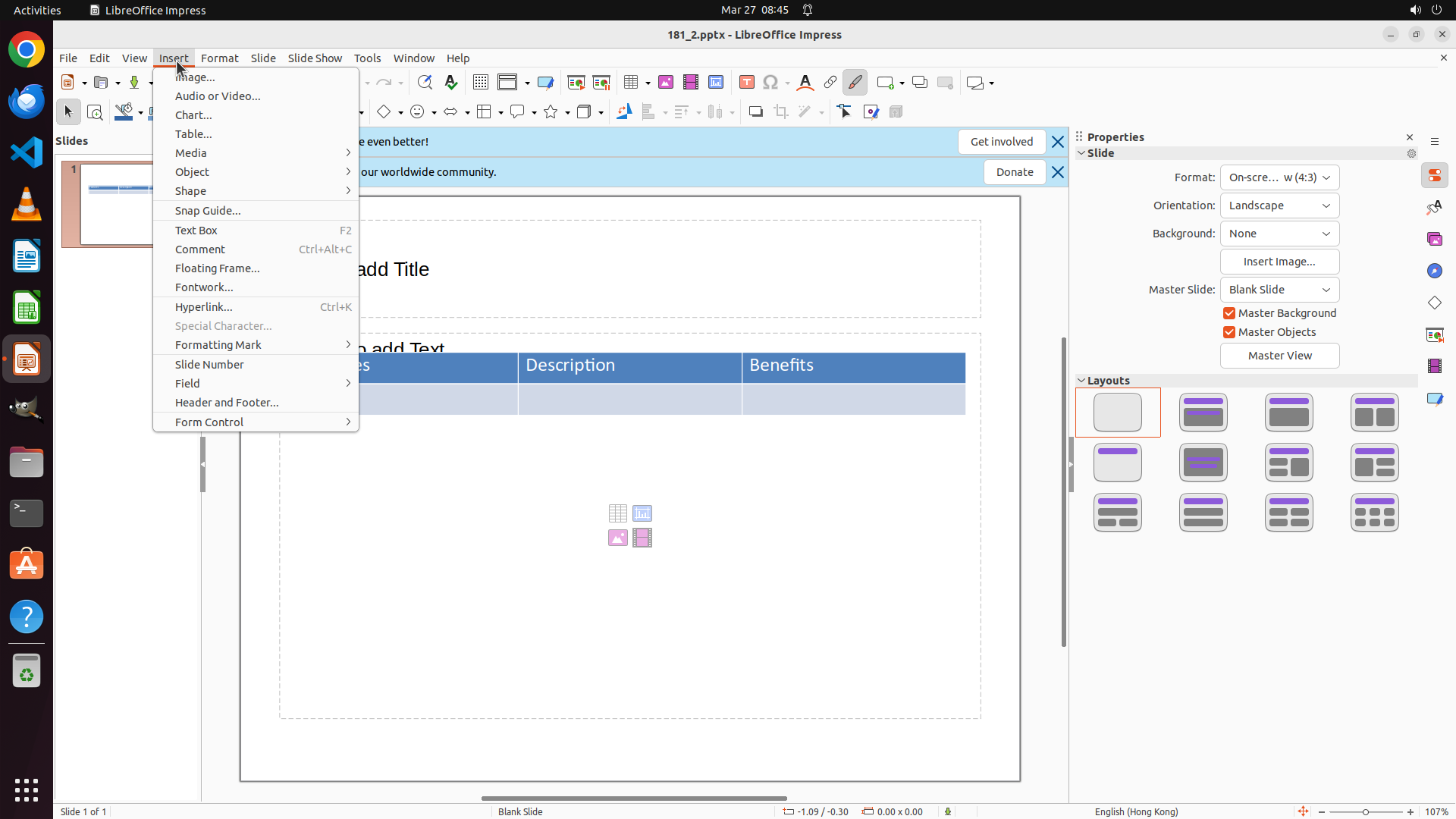} &
\includegraphics[width=0.30\linewidth]{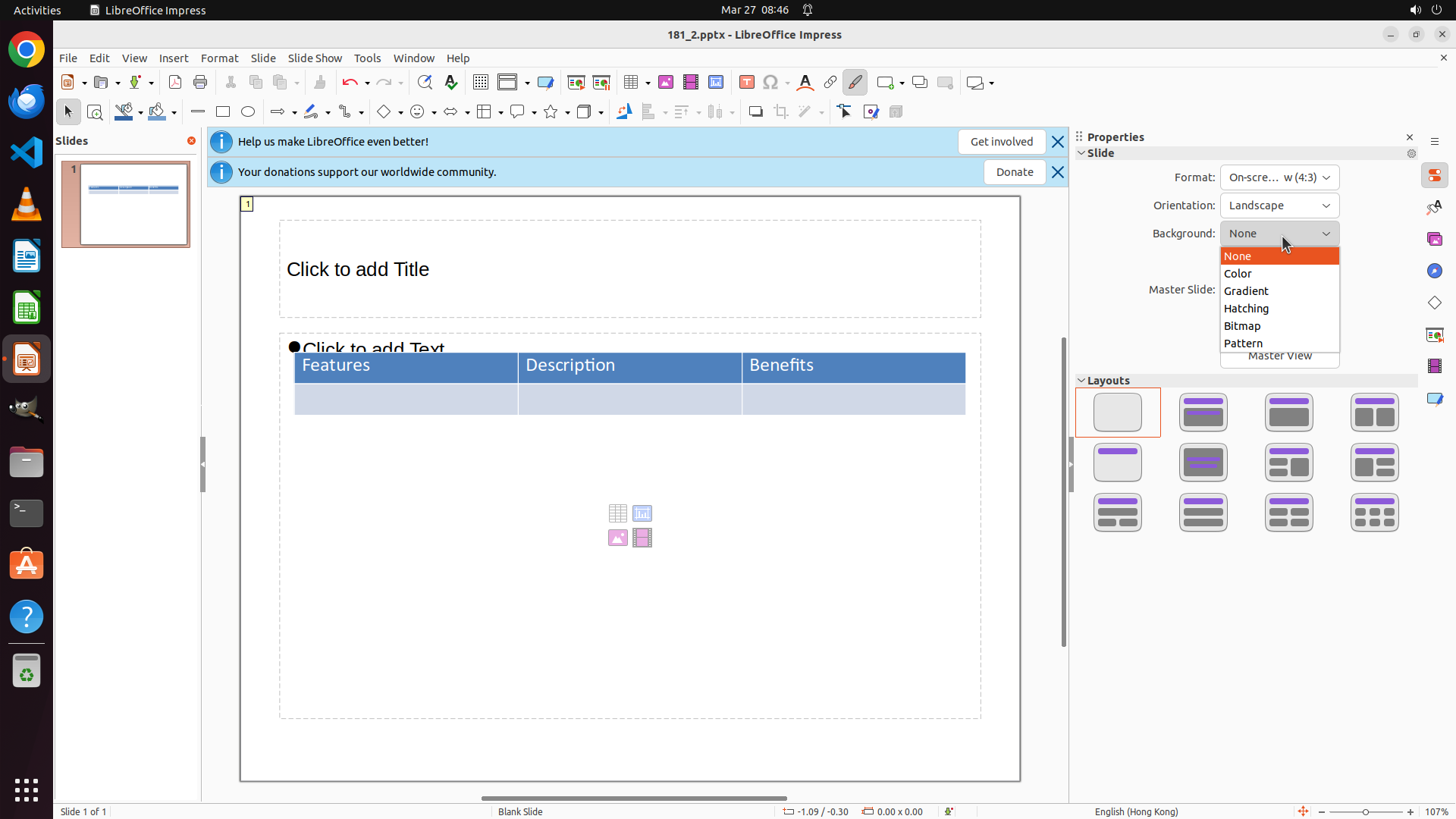} \\[-2pt]
{\scriptsize (a) Initial: slide in Impress} &
{\scriptsize (b) Open Insert menu (Comment option)} &
{\scriptsize (c) Set background to purple} \\[6pt]
\includegraphics[width=0.30\linewidth]{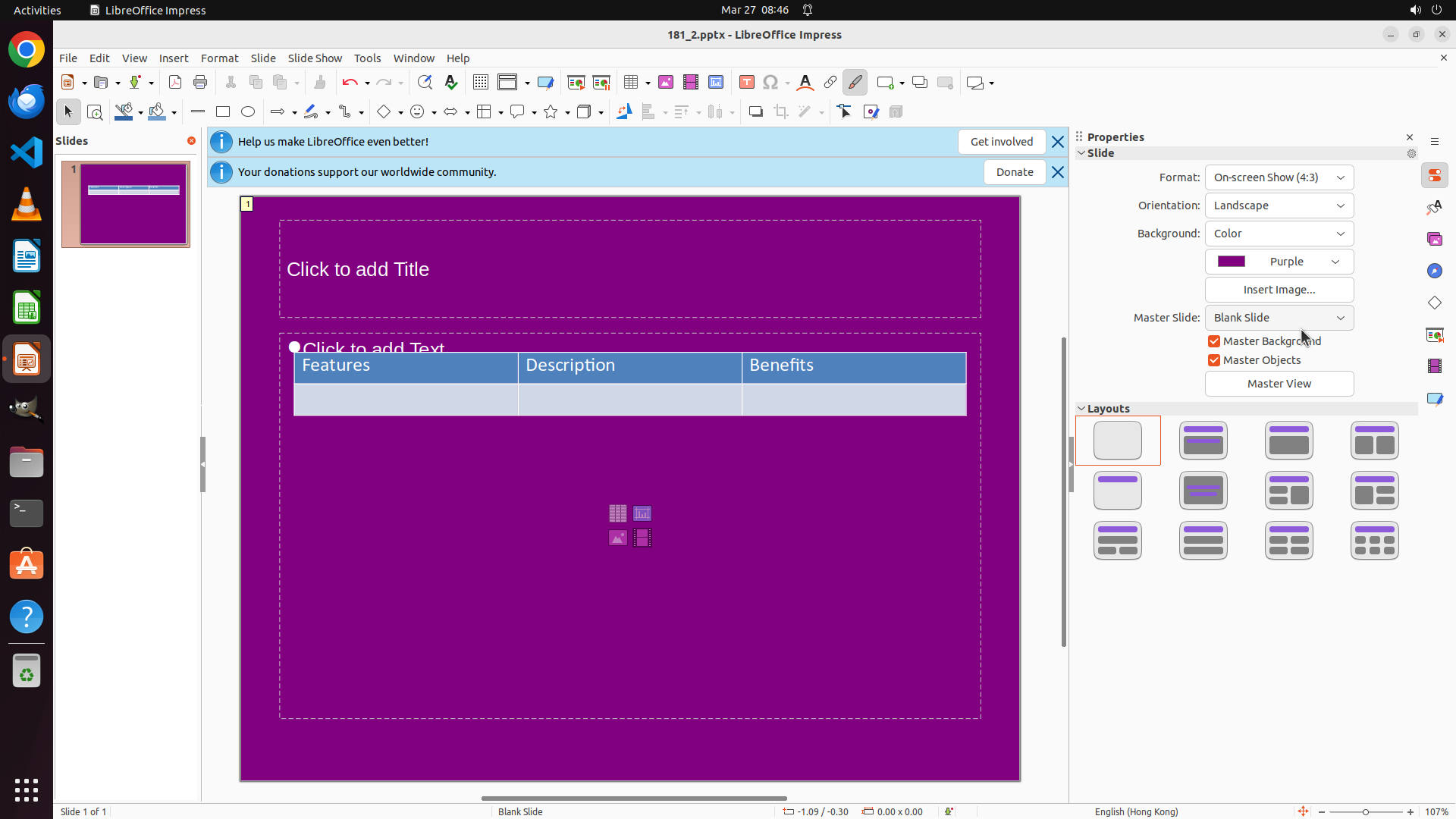} &
& \\[-2pt]
{\scriptsize (d) Purple background applied, calls \texttt{done()}} & & \\
\end{tabular}

\begin{tcolorbox}[colback=red!5!white,colframe=red!50!black,title={\small User Simulator Feedback}]
\small You added a comment on the slide instead of a slide note. The expected change is that the slide's Notes (speaker notes) contain exactly ``APP'' (case-sensitive). The purple background color looks correct. Click into the Notes pane below the slide and type ``APP'' there.
\end{tcolorbox}

\noindent\textbf{After feedback (succeeded):}
The user simulator disambiguated ``note'' as speaker notes. The agent clicked into the Notes pane at the bottom of the Impress window and typed ``APP.''

\vspace{0.3em}
\noindent\begin{tabular}{@{}ccc@{}}
\includegraphics[width=0.30\linewidth]{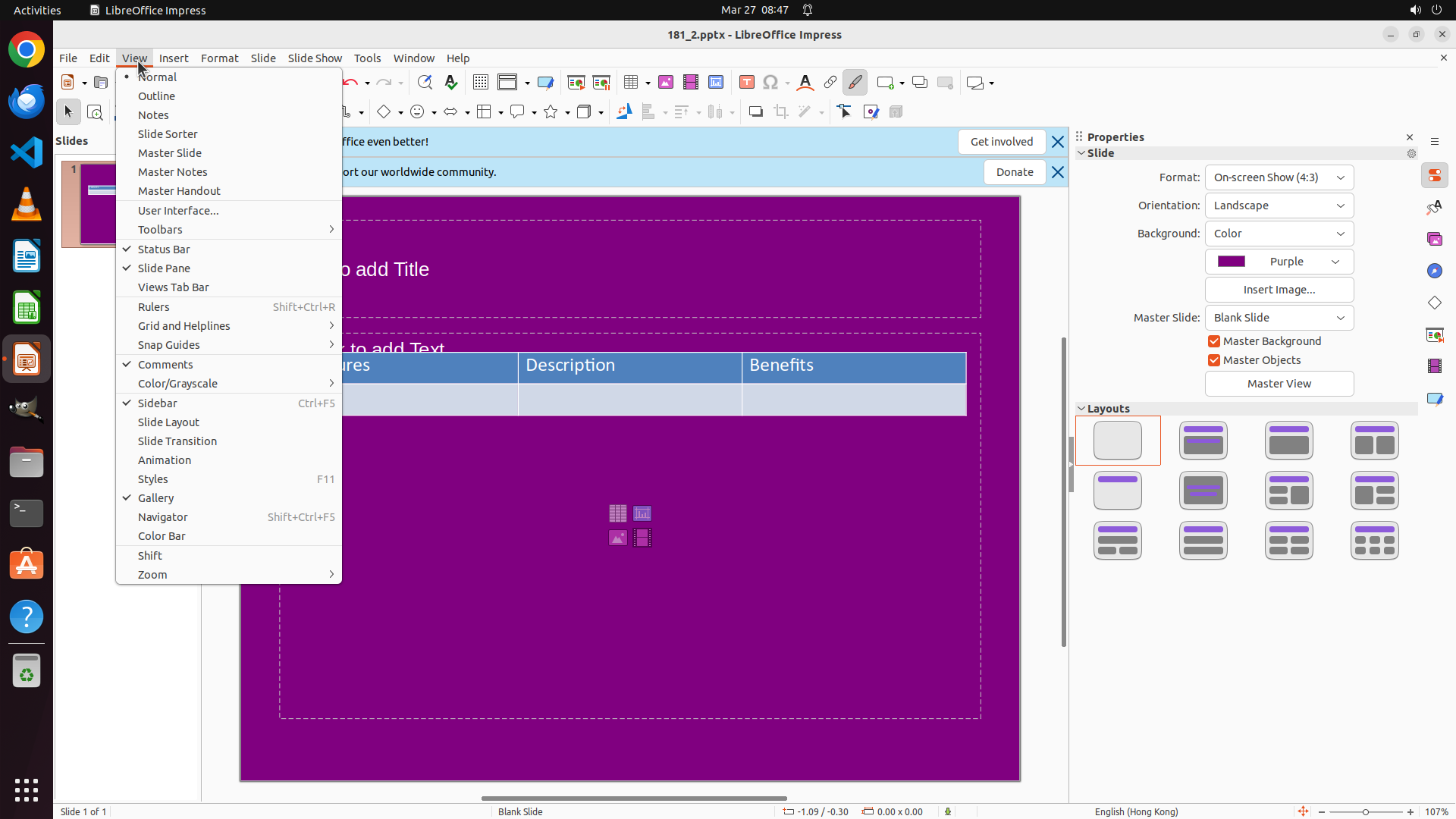} &
\includegraphics[width=0.30\linewidth]{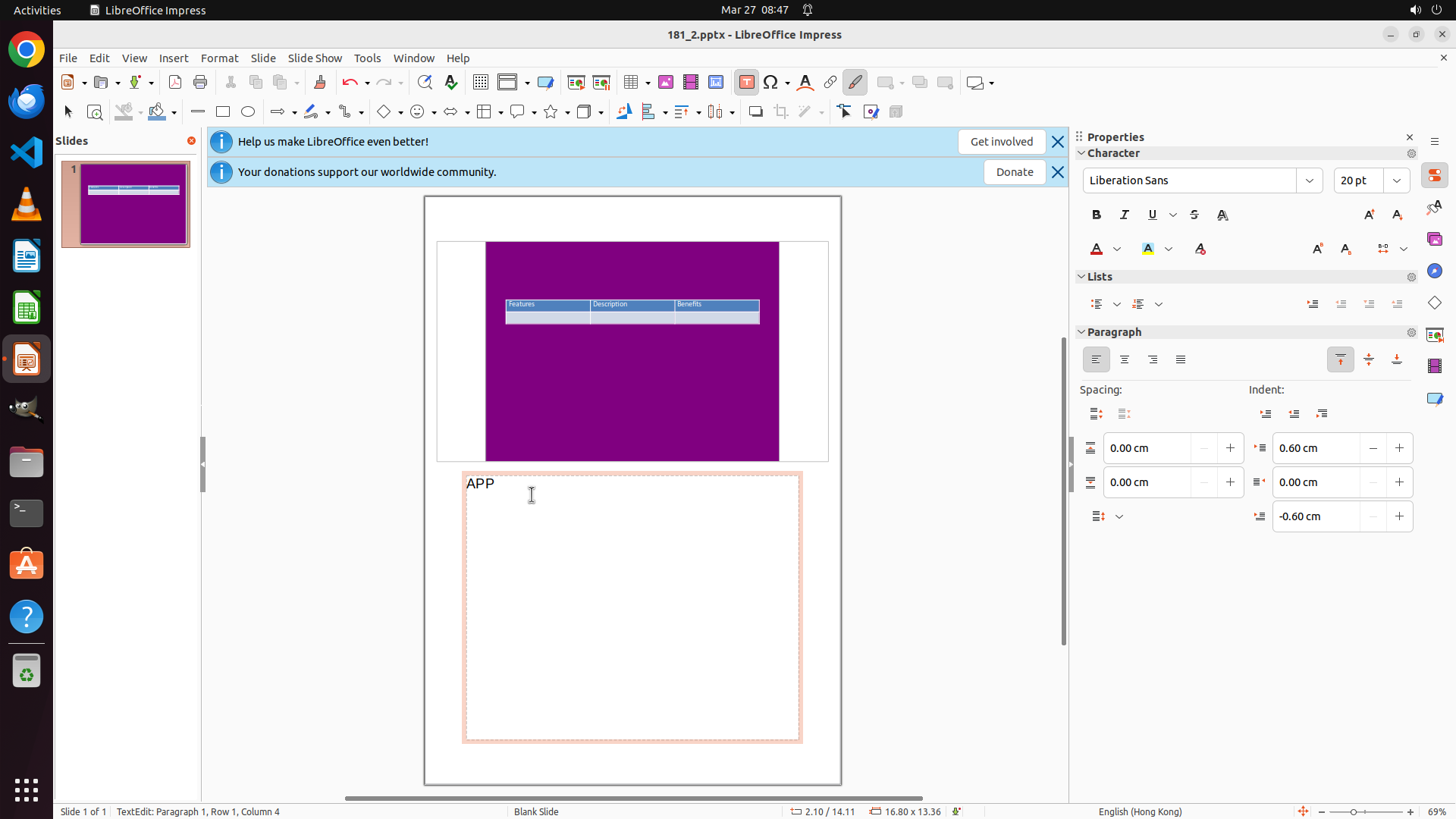} &
\includegraphics[width=0.30\linewidth]{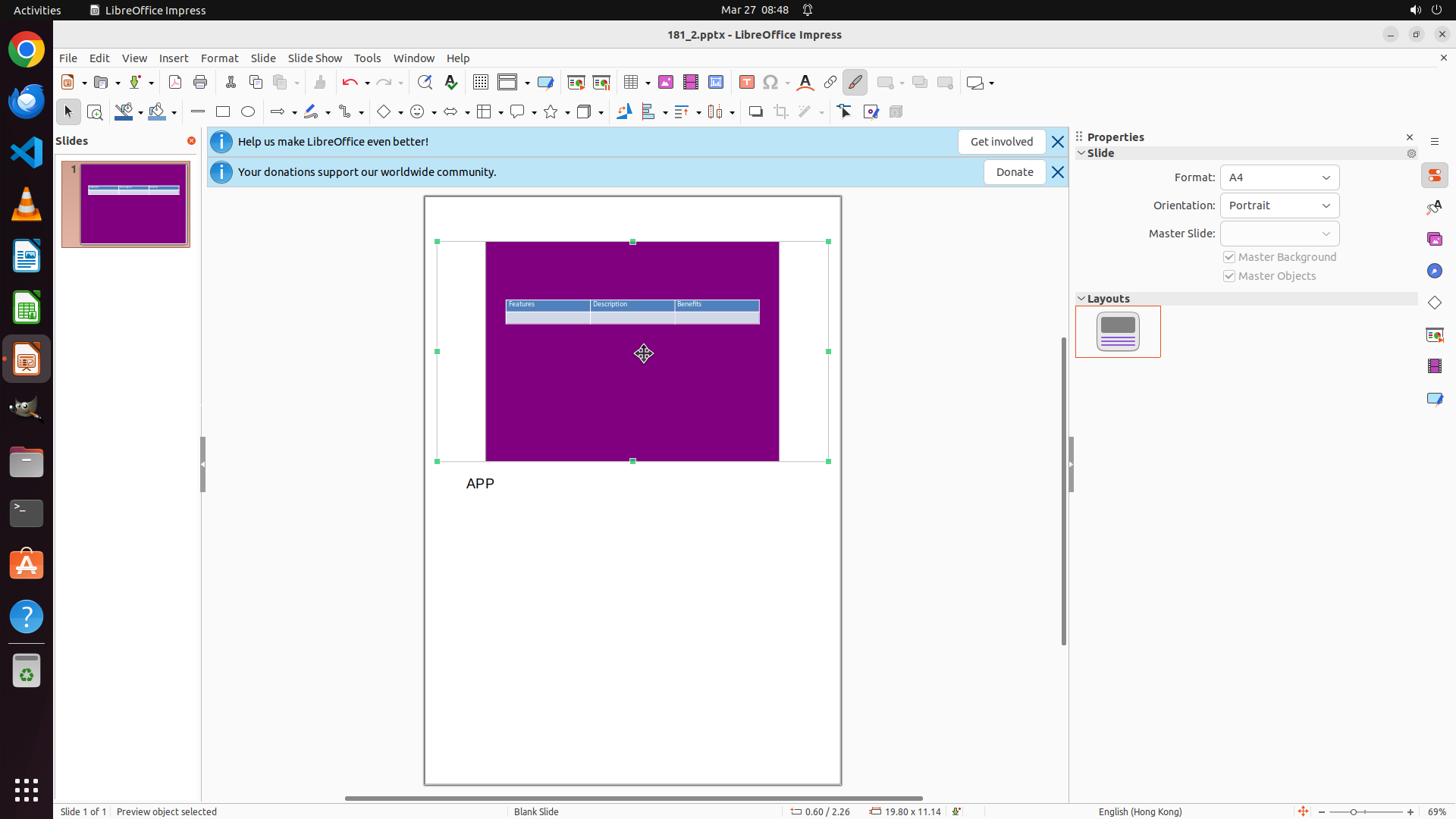} \\[-2pt]
{\scriptsize (d) Open View menu $\to$ Notes} &
{\scriptsize (e) Type ``APP'' in Notes pane} &
{\scriptsize (f) Notes pane shows ``APP'', calls \texttt{done()}} \\
\end{tabular}

\clearpage
\section{Clarification Details}
\label{app:clarification_details}
To study the effect of reducing instruction ambiguity, we construct clarified versions of task instructions by making minimal edits that align the instruction with the evaluator’s success criteria. Given the original instruction, the current environment state (screenshot), the task configuration (evaluation JSON), and the associated evaluator implementation, we prompt an LLM (GPT-5) to identify aspects of the evaluator that are not explicitly specified in the instruction and to produce a minimally clarified version. The prompt enforces that clarifications preserve the original intent and natural, human-like tone, while only adding details required by the evaluator (e.g., exact file names, formats, or locations). The full prompt is provided in Appendix~\ref{prompt:ic:clarification}.

To ensure that clarification improves alignment with evaluation criteria without introducing unintended artifacts, we perform a verification and human correction procedure. For each task, we evaluate both the original and clarified instructions using multiple independent runs (three runs each with GPT-5 and Claude Sonnet 4.6) and compare the resulting success rates. We then use a modified version of Behavior Judge \citep{agents3} to compare trajectories from both settings (Appendix~\ref{prompt:ic:trajectory_analysis}), conditioned on the evaluator implementation, and identify whether changes in success rate are due to improved alignment, random variation, or issues introduced by clarification (e.g., making a task impossible or overly trivial) (Appendix~\ref{prompt:ic:trajectory_analysis_input}). We flag cases with substantial performance shifts (e.g., consistently solved to never solved and vice versa). These cases are then manually reviewed, using the judge’s analysis as guidance, the raw trajectories, and the evaluator requirements to guide minimal edits that correct problematic instructions. In total, we manually corrected 25 out of 361 tasks, including 20 cases where clarification introduced impossible constraints and 5 where it made the task overly trivial. This procedure ensures that the clarified instructions better reflect evaluator requirements while preserving task difficulty. Examples of corrections are shown in Appendix~\ref{app:clarification:qualitative}.

\section{Environment Perturbation Details}
\label{app:perturbation_details}
\begin{table}[h]
    \centering
    \caption{Perturbation sets. Each row describes one modified desktop property. All modifications are purely cosmetic and do not affect application functionality or task correctness.}
    \label{tab:perturbation_details}
    \begin{tabular}{lccc}
    \toprule
    \textbf{Property} & \textbf{Default} & \textbf{Set 1} & \textbf{Set 2} \\
    \midrule
    Wallpaper & Jammy Jellyfish & Jammy Jellyfish & Optical Fibers \\
    Cursor size & 24\,px & 48\,px & 16\,px \\
    Dock position & Left & Bottom & Right \\
    Icon theme & Yaru & Adwaita & Humanity \\
    Timezone & America/Los\_Angeles & Asia/Tokyo & Europe/London \\
    \bottomrule
    \end{tabular}
\end{table}

\end{document}